# High-Definition Map Generation Technologies for Autonomous Driving

Zhibin Bao, Sabir Hossain, Haoxiang Lang, Xianke Lin*

*Abstract*— Autonomous driving has been among the most popular and challenging topics in the past few years. On the road to achieving full autonomy, researchers have utilized various sensors, such as LiDAR, camera, Inertial Measurement Unit (IMU), and GPS, and developed intelligent algorithms for autonomous driving applications such as object detection, object segmentation, obstacle avoidance, and path planning. High-definition (HD) maps have drawn lots of attention in recent years. Because of the high precision and informative level of HD maps in localization, it has immediately become one of the critical components of autonomous driving. From big organizations like Baidu Apollo, NVIDIA, and TomTom to individual researchers, researchers have created HD maps for different scenes and purposes for autonomous driving. It is necessary to review the state-of-the-art methods for HD map generation. This paper reviews recent HD map generation technologies that leverage both 2D and 3D map generation. This review introduces the concept of HD maps and their usefulness in autonomous driving and gives a detailed overview of HD map generation techniques. We will also discuss the limitations of the current HD map generation technologies to motivate future research.

## I. INTRODUCTION

THE "high-definition map" concept was first introduced in the Mercedes-Benz research in 2010 and later contributed to the Bertha Drive Project [1] in 2013. In the Bertha Drive Project, a Mercedes-Benz S500 completed the Bertha Benz Memorial Route in a fully autonomous mode, utilizing a highly precise and informative 3D road map, which was later named as "High Definition (HD) Live Map" by a participating mapping company called HERE [2]. The HD map contains all critical static properties (for example: roads, buildings, traffic lights, and road markings) of the road/environment necessary for autonomous driving, including the object that cannot be appropriately detected by sensors due to occlusion. HD maps for autonomous driving have been known for their high precision and rich geometric and semantic information in recent years. It tightly connects with vehicle localization functionality and constantly interacts with different sensors, including lidar, radar, and camera, to construct the perception module of the autonomous system. This interaction ultimately supports the mission and motion planning of the autonomous vehicle [3], as shown in Fig. 1.

TABLE I
EXAMPLES OF THE THREE-LAYER STRUCTURED HD MAP

| Layer Number | TomTom | HERE | Lanelet (Bertha Drive) |
|---|---|---|---|
| 1 | Navigation Data | HD Road | Road Network (OpenStreetMap) |
| 2 | Planning Data | HD Lanes | Lane Level Map |
| 3 | Road DNA | HD Localization | Landmarks/Road Marking Map |

There is no unique standard HD map structure in the autonomous driving market. However, there are some commonly adopted structures for HD maps on the market, such as Navigation Data Standard (NDS), Dynamic Map Platform (DMP), HERE HD Live Map, and TomTom [2]. Most structures share a similar three-layer data structure. Table I shows the three-layer structured HD map defined by TomTom, HERE, and Lanelet (Bertha Drive). This paper will adopt HERE's terminology to refer to the three layers, as shown in Fig. 2.

The first layer, Road Model, defines the road characteristics, such as topology, the direction of travel, elevation, slope/ramps, rules, curbs/boundaries, and intersections. It is used for navigation. The second layer, Lane Model, defines the lane level features, such as road types, lines, road widths, stop areas, and speed limits. This layer works as a perception module for autonomous driving to make decisions based on the real-time traffic or environment. As the name implies, the last layer, Localization Model, localizes the automated vehicle in the HD map. This layer contains roadside furniture, such as buildings, traffic signals, signs, and road surface markings. Those features help the automated vehicle quickly locate itself, especially in the urban area where it is rich in features. The HD maps

*This work was supported by the Dr Xianke Lin's startup funding, CANADA." (Corresponding author: X. Lin).

The authors are with Department of Automotive and Mechatronics Engineering, Ontario Tech University, Oshawa, ON L1G 0C5, Canada. (e-mail: zhibin.bao@ontariotechu.net, sabir.hossain@ontariotechu.net, Haoxiang.Lang@ontariotechu.ca, xiankelin@ieee.org*).

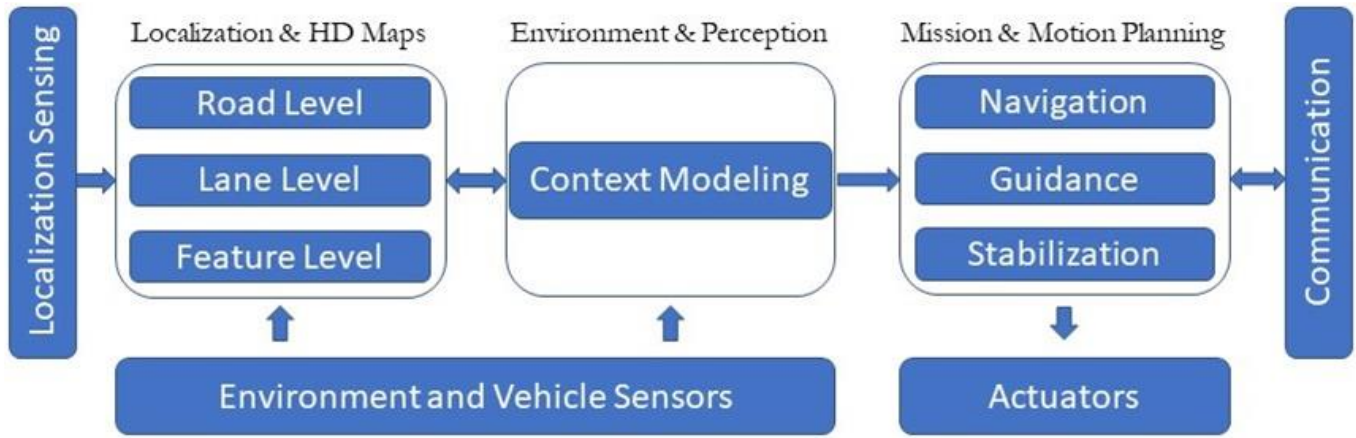

Fig. 1. Autonomous System Architecture: HD map contains static information and properties about the road/environment, including objects that cannot be detected by the perception module due to occlusion. It also localizes the ego vehicle based on the road features. The environment & perception module provides real-time environmental information around the ego vehicle. HD map and the perception module work together to ultimately support the mission and motion planning module, including navigation, motion guidance, and stability control.

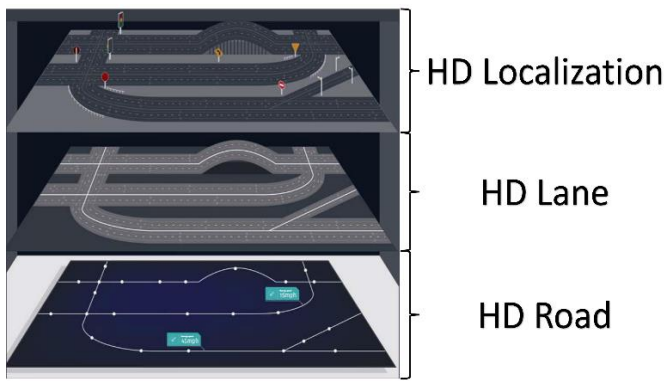

Fig. 2. HD map structure defined by HERE: HD road consists of topology, the direction of travel, intersections, slope, ramps, rules, boundaries, and tunnels. HD lanes consist of lane level features, such as boundaries, types, lines, and widths. HD localization consists of road furniture, such as traffic lights and traffic signs.

constructed by the above organizations are precise and constantly updated. However, they are only for commercialized purposes and not open source. It is hard for individual researchers to construct HD maps using the above structures. Thus, this paper will review uncommercialized HD map generation methods that can potentially help researchers create their personalized HD maps and develop new HD map generation methodologies.

The structure of this paper is organized as follows: Section II reviews the recent data collection methods for HD maps. Section III reviews and compares the point cloud map generation methods. Section IV reviews the recent feature extraction methods for HD maps, including road/lane network extraction, road marking extraction, and pole-like object extraction. Section V introduces commonly adopted frameworks for HD maps. Section VI will discuss the limitations of the current HD maps generation methods and bring out some open challenges for HD maps. Lastly, the conclusion is given in Section VII.

## II. DATA COLLECTION FOR HD MAPS

Data sourcing/collection is the first step in generating HD maps. Data collection is done using a mobile mapping system (MMS). An MMS is a mobile vehicle fitted with mapping sensors including GNSS (Global Navigation Satellite System), IMU, LiDAR (light detection and ranging), camera, and radar to collect geospatial data. Commercialized HD map providers adopt crowdsourcing to collect data for building and maintaining their HD maps. Level5 collaborated with Lyft to send 20 autonomous vehicles along a fixed route in Palo Alto, California, to collect the dataset consisting of 170,000 scenes, an HD semantic map with 15242 labelled elements, and an HD aerial view over the area [4]. TomTom collected data through multi-source approaches, including survey vehicles, GPS traces, community input, governmental sources, and vehicle sensor data [5]. HERE utilized over 400 mapping vehicles worldwide, governmental data, satellite imagery, and community input to constantly get updated road information [6]. Data collection via crowdsourcing can collect a vast amount of up-to-date road/traffic data within a small amount of time. The crowdsourced data also contains different environments, including cities, towns, and rural areas. However, this method is not an optimal solution for individual researchers due to the high cost of multiple mobile mapping systems and the time consumption of data collection. Individual researchers also utilize the MMS to collect data. Instead of collecting data for different types of environments around the world, they focus on a much smaller scaled area, such as a city, a university campus, or a residential area. The collected data types are also more specified for research purposes. Additionally, there are also plenty of open-sourced data such as satellite images, KITTI dataset [7], Level5 Lyft Dataset [4], and nuScenes Dataset [8] for researchers to do testing and generate HD maps. Those datasets contain 2D and 3D real-world traffic data, including images, 3D point clouds, and IMU/GPS data, which have already been organized and labelled. The data collection methods and comparison are summarized in Table II.

TABLE II
DATA COLLECTION METHODS COMPARISON

| Comparison | Data Collection Methods | | |
|---|---|---|---|
| | *Crowdsourcing* | *Open-sourced dataset* | *Self-collected dataset* |
| Advantages | High efficiency, Up-to-date data, Various types of environments | High quality, Timesaving in data collection, Data is organized and does not require post-processing | Low cost compared to crowdsourcing, Customizable data collection (specific scene/environment, sensor types, data size) |
| Disadvantages | High cost, Post-processing of raw data is time-consuming and computationally expensive | Data may not be up-to-date, limited types of environments or scenarios | Data collection for large-scale maps is time-consuming, data may not be up-to-date, small mapping coverage |

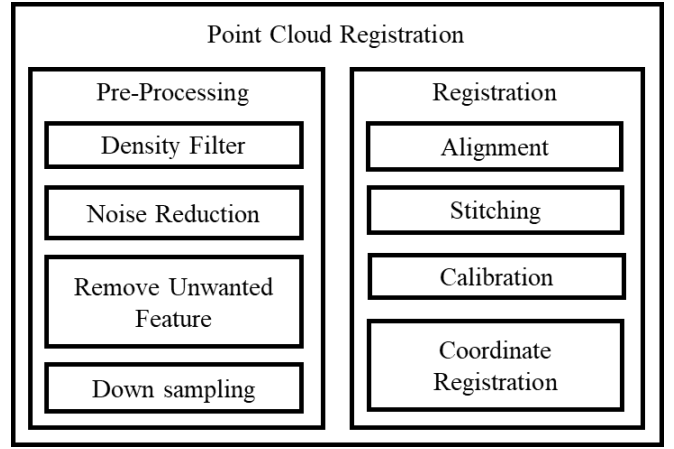

Fig. 3. Common Multi-step process of Point Cloud Registration

## III. POINT CLOUD MAP GENERATION

Once initial sensor data is collected, it is usually fused and sorted to generate the initial map, mainly used for accurate localization. Initial mapping is primarily generated using a 3D laser sensor; however, it can be fused with other sensors such as IMU [9]–[11], GPS [12], odometry [13] and visual odometry [14] for more accurate state estimation in the HD map. INS and GPS sensors provide orientation and position information that updates the mapping location within cm-accuracy. These point cloud maps have been highly accurate and later assist the ground vehicle in precise maneuver and localization at centimeter-level in the 3D space. Later, a vector map is created from the PCL map after the point cloud registration is obtained from the mapping. Point cloud registration is called a multi-step process (shown in the Fig. 3) of aligning several overlapped point clouds to generate a detailed and accurate map. The vector map holds information related to the lane, sidewalk, crossings, roads, intersections, traffic signs, and traffic lights. This critical feature was later utilized to detect traffic signs and lights, route planning, global planning and local path planning. Undoubtedly, map generation is an integral part of high-definition map generation. It can be defined as an HD map's base geometry map layer.

### A. Mapping Techniques

Map generation techniques can be classified into online and offline mapping. Offline mapping data is collected all in a central location. The data are using satellite information or data stored from LiDAR and camera. The map is built offline after data is collected. On the other hand, map generation happens onboard using lightweight modules in online mapping. Apart from map formation type, mapping techniques can be classified by using sensors or how the sensors are fused. The following mapping techniques require laser-based sensors since they show promising accuracy at long distances.

All promising mapping technology currently uses the laser as a major sensor for mapping and completing the high-definition map. On the other hand, there are approaches which only use vision sensors to construct the point cloud map. There is a point cloud registrations technique which is developed for 3D model generation. However, the following methods are classified on the basis of mapping that will support HD maps.

*1) Segmentation-based Point Cloud Registration*

SegMap [15] is a mapping solution based on the extraction of segmented features in the point clouds. The approach generates the point cloud map by reconstructing the local features to be discriminative. The trajectory result shows an enhanced performance when combined with LOAM (Laser Odometry and Mapping) than only the LOAM framework [16]. The constructed map displays 6% accuracy in recall and 50% decrement in Odometry drift. Therefore, there is an improvement in localization due to data-driven segment descriptor since it provides less coarse data. A simple, fully connected network is trained for the SegMap descriptor, and later map is reconstructed from the semantic extraction.

Similarly, a two-stage algorithm is used to improve the mapping errors. It is performed using a segment matching algorithm integrated with LiDAR only algorithm. Also, RANSAC-based geometrical enhancement is introduced to reduce the false match between the generated map and online mapping [17]. Such LiDAR only methods are discussed in the next section.

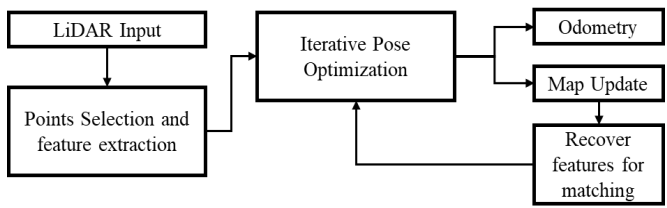

Fig. 4. Common Mapping Workflow

*2) Only LiDAR-based Point Cloud Mapping*

Superior precision and efficiency are already achieved from LiDARs with small FoV (field of view) and irregular samplings by improving the existing point selection method and iterative pose optimization method of LOAM [18]. The overall mapping architecture is shown in Fig 4. A fast loop closure technique is introduced to fix the long-term shift in LiDAR odometry and mapping [19]. On the other hand, a decentralized multi-LiDAR platform of small FoV is used for robust mapping using an extended Kalman filter [20]. Additionally, there is a technique where LiDARs are equipped in the robot at different heights to generate the point cloud [21].

*3) Odometry Fused Point Cloud Registration*

Fusing odometry comes in handy when GPS is not available or gets disconnected, mostly indoors. The iterative closest point (ICP) method uses 6-DOF information to match the closest geometry in the given point cloud. The main drawback of this is that it is stuck at local minima and requires a perfect starting point, causing an increase in error and misalignment with the real environment [22]. NDTMap [23], [24] generation is a continuous and differentiable probability density transformed from point cloud [25], [26]. The probability density of NDTMap contains a set of normal distributions. It is a voxel grid where each point is assigned to a voxel-based on its coordinates. Points cloud is divided into voxel cloud, and the merged voxel is later filtered for noise reduction in the map and less computation. So, the overview of NDT mapping follows the steps below –
- Building a voxel grid from the point cloud input data
- Estimating the initial guess
- Optimization of initial guess
- State estimation from the translational change between NDT estimate and an initial guess. Velocity and acceleration calculation from the positional derivations.

If odometry is not used in the initial guess, the state estimation is derived from each NDT update. The initial guess comes from the velocity and acceleration update based on the motion model. When odometry is introduced, the position update is based on odometry data; specifically the velocity model and orientation update.

*4) GPS Fused Point Cloud Registration*

Absolute position is included from GNSS as a constraint in the graph-based mapping to unify the point cloud data with the coordinate system [12]. Therefore, voxels in the point cloud are tagged with absolute 3D coronadite information. LiDAR-based odometry is also used in LIO-SAM for accurate pose estimation and map building [13].

*5) INS Fused Point Cloud Registration*

Without using any sensors, vehicle state and yaw are calculated from each NDT update. The motion model-based initial guess is derived using velocities and acceleration. IMU provides translational updates to the quadric model and orientation updates. NDT mapping technique by Autoware [27] also provides IMU and odometry fusion for mapping. Similarly, the DLIO method [28] enables precise mapping and high-rate state estimation by using both loosely-coupled fusion and pose graph optimization. IMU is integrated to enhance reliability by feeding IMU biases to correct subsequent linear acceleration and angular velocity values. FAST-LIO [10] and FAST-LIO2 [11] are LiDAR-inertial odometry systems for fast and accurate mapping. IMU is fused with LiDAR feature points in this system using a tightly-coupled iterated EKF (Extended Kalman filter). FAST-LIO2 uses a novel technique, incremental kd-Tree, that provides an incremental update and dynamic rebalancing sustaining the map.

*6) Visual Sensor Fused Point Cloud Registration*

R2-LIVE [29] and R3-LIVE [30] algorithms use the fusion of Laser, INS and the visual sensor for accurate mapping and state estimation. R2-LIVE confirms accurate state estimation using Kalman-filter-based iterated odometry and factor graph optimization. R3-LIVE is the combination of two separate modules: LiDAR-IMU odometry and visual-IMU odometry. The global map achieves precise geometric measurement from LiDAR and IMU. The visual sensor fused with IMU projects the map texture into the global map. Similar two submodules, LIO and VIO are used for robust and accurate mapping in FAST-LIVO [14] also. Similarly, LVI-SAM is designed using two sub-modules similar to R3-LIVE. According to LVI-SAM [31], a visual inertial system leverages LiDAR inertial calculation to assist the initialization. Visual sensors provide depth information to improve the accuracy of the visual inertial system.

Fig. 5 shows generated map using existing mapping algorithms. There are available techniques where multiple sensors are fused to create a complete map. Visual odometry (IMU and camera), GPS, and LiDAR data are combined into a supernode to get an optimized map [32]. Fig 6 shows the trajectory path obtained from the online mapping using different approaches. Fig 6(a) is the complete path of the mapping sensor data (Ontario Tech Campus). Fig 6(a) shows the complete odometry data from the recorded data. Fig 6(b) and fig 6(c) are the enlarged versions of the full trajectory path. This ground-truth path is obtained from the fusion of RTK-GPS and IMU data. These fractions show that R3-LIVE follows the ground-truth path, which is RTK-GPS odometry.

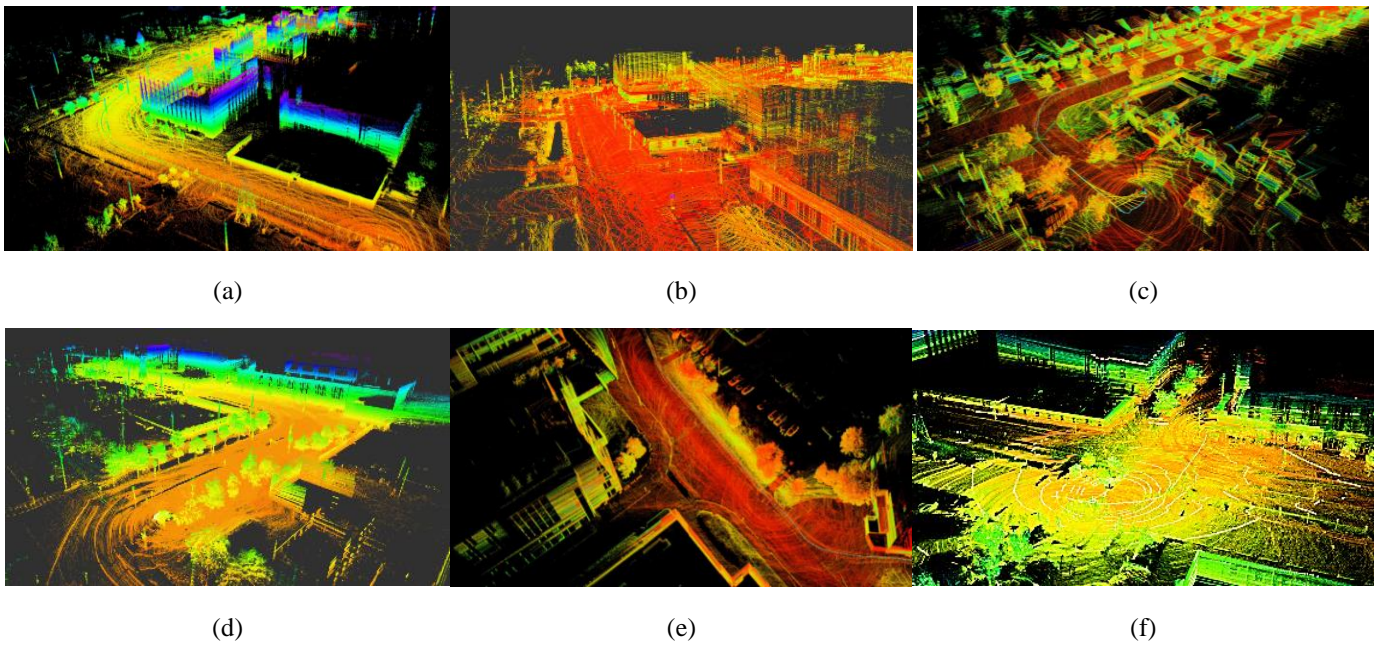

Fig. 5. Mapping Visualization; (a) LeGO-LOAM [9], (b) NDT Mapping [24] (Autoware [27]), (c) LIO-SAM [13], (d) FAST-LIO [10], (e) LVI-SAM [31], (f) R3-LIVE [30]

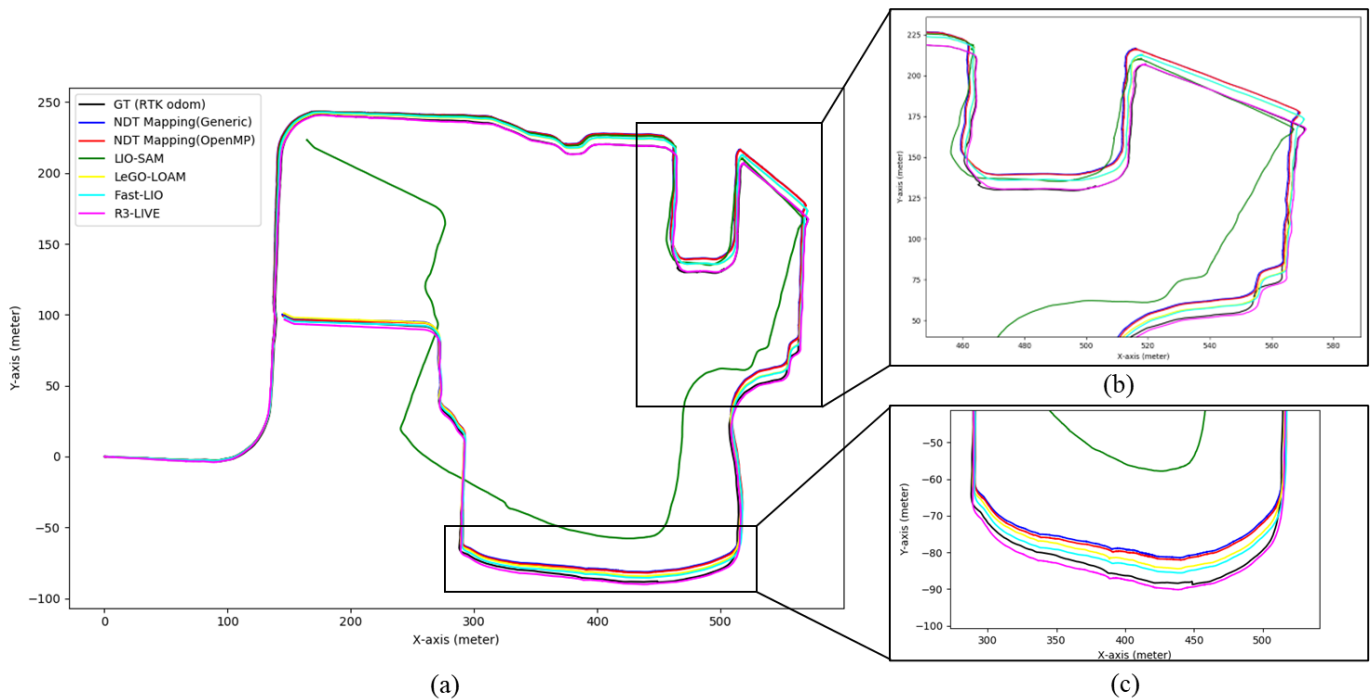

Fig. 6. Odometry Path Plot from Different Mapping Algorithms

On the other hand, LIO-SAM [13] drifted in the middle from the original course (GPS odometry not used). For NDT Mapping, generic PCL transformation and OpenMP method are used. The odometry value used for plotting was the after mapped (For example, LeGO-LOAM [9] odometry was plotted after the additional step of performing map matching).

## IV. FEATURE EXTRACTION METHODS FOR HD MAP

To have the ego vehicle locate itself and follow the motion and mission plans, feature extractions such as road/lane extraction, road marking extraction, and pole-like object extraction are necessary. Feature extractions were traditionally done by labor, which was costly, time-consuming, and low precision. In recent years, machine learning assisted HD map generation techniques have been developed and widely used to increase feature extraction precision and reduce the amount of manual work.

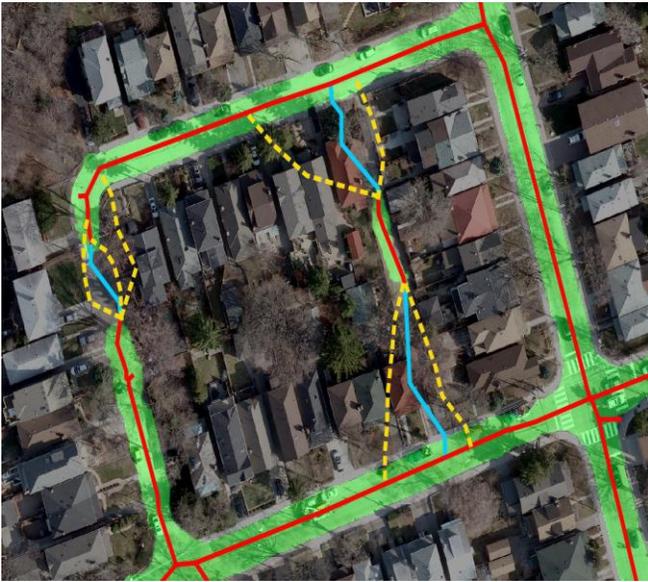

Fig. 7. Road segmentation is highlighted in green, the red lines are the extracted road centerlines, yellow dash lines show the potential connections for the road, and blue lines are the potential lines selected by the A* algorithm [36].

Machine learning-assisted HD map generation utilizes the human-in-the-loop (HITL) technique, which involves the interaction of both human and machine [33]–[35]. Humans do data labeling, and the labeled data is trained using supervised learning. Results with high precision/confidence scores will be saved to the HD map, and results with low precision/confidence scores will be examined by humans and sent back to the algorithm for retraining. Machine learning has been widely adopted in extracting road/lane networks, road markings, and traffic lights.

### A. Road Network Extraction

#### 1) Road Extraction on 2D Aerial Images

Road maps/networks are essential for an autonomous driving system to localize the ego vehicle and plan routing. Extracting road maps from aerial images is also appealing since aerial photos have extensive coverage of maps, usually at the city scale, and are constantly updated through satellites. However, manually creating road maps from aerial images is labour-intensive and time-consuming. It also does not guarantee a precise road map due to human error. Thus, methods that can automate the road map extracting process is in demand.

Automatic road network extractions of 2D aerial images can be divided into three different methods: segmentation-based, iterative graph growing, and graph-generation methods.

*a) Segmentation-based Method*

Segmentation-based methods predict the segmentation probabilistic map from aerial images and refine the segmentation predictions and extract the graph through post-processing.

Mattyus et al. proposed an approach that directly estimates the road topology and extracts the road network from aerial images [36]. In their approach, named DeepRoadMapper, they first used a variant of ResNet [37] to segment the aerial images into interest categories. Then, they filtered the road class with a threshold of 0.5 probability using the softmax activation function and extracted the centerlines of the road using shinning [38]. To alleviate the discontinuity issue of the road segmentation, they connected the endpoint of the discontinued road to other roads' endpoints within a specific range. The connections are considered potential roads, and the $A^*$ algorithm [39] is applied here to select the shortest connections as the road for discontinuity, see Fig. 7.

By evaluating their approach on the TorontoCity dataset [40] and comparing the results with [41], OpenStreetMap, and the ground truth map, they showed significant improvements over the state-of-the-art at the year of publication. Besides the improvements, it is noticeable that the heuristics ($A^*$ algorithm) is not an optimal solution when the road or surrounding environment complexity increases, such as occlusion.

To enhance the segmentation-based road network extraction performance and solve the road network disconnectivity issue in [36], [42] proposed the *Orientation Learning* and *Connectivity Refinement* approaches. The proposed approaches fix the road network disconnectivity issue by predicting the orientation and segmentation of the road network and correcting the segmentation results using an n-stacked multi-branch CNN. This method is further evaluated on the SpaceNet [43] and DeepGlobe [44] dataset and compared with DeepRoadMapper and other state-of-the-art methods [45]–[48] to show its state-of-the-art results. The evaluation results are shown in Table III. The bold highlighted values represent the best results. Based on the comparison in Table III, OrientationiRefine has the best state-of-the-art result.

Furthermore, Ghandorh et al. refined the segmented road networks from satellite images by adding an edge detection algorithm to the segmentation-based method [49]. The proposed approach used the encoder-decoder architecture along with dilated convolutional layers [50] and the attention mechanism [51]–[54] to give the network ability to segment large-scale objects and focus more on the important features. The segmented road networks were then further refined by feeding them into the edge detection algorithm.

*b) Iterative Graph Growing Method*

Iterative graph growing methods generate the road network from 2D aerial images by first selecting several vertices of the road network. Then, the road is generated vertex by vertex until the whole road network is created.

TABLE III
COMPARISON OF THE STATE-OF-THE-ART ROAD EXTRACTION METHODS ON SPACENET AND DEEPGLOBE DATASET.
$IOU^R$ AND $IOU^A$ REFERS TO RELAXED AND ACCURATE ROAD $IOU$. $APLS$ REFERS TO AVERAGE PATH LENGTH SIMILARITY [41].

| Method | SpaceNet | | | | | | DeepGlobe | | | | | |
|---|---|---|---|---|---|---|---|---|---|---|---|---|
| | Precision | Recall | F1 | $IoU^r$ | $IoU^a$ | APLS | Precision | Recall | F1 | $IoU^r$ | $IoU^a$ | APLS |
| DeepRoadMapper (segmentation) [36] | 60.61 | 60.80 | 60.71 | 43.58 | 59.99 | 54.25 | 79.82 | 80.31 | 80.07 | 66.76 | 62.58 | 65.56 |
| DeepRoadMapper (full) [36] | 57.57 | 58.29 | 57.93 | 40.77 | N/A | 50.59 | 77.15 | 77.48 | 77.32 | 63.02 | N/A | 61.66 |
| Topology Loss (with BCE) [45] | 50.35 | 50.32 | 50.34 | 33.63 | 56.29 | 49.00 | 76.69 | 75.76 | 76.22 | 61.58 | 64.95 | 56.91 |
| Topology Loss (with Soft IoU) [45] | 52.94 | 52.86 | 52.90 | 35.96 | 57.69 | 51.99 | 79.63 | 79.88 | 79.75 | 66.32 | 64.94 | 65.96 |
| LinkNet34 [47] | 61.30 | 61.45 | 61.39 | 44.27 | 60.33 | 55.69 | 78.34 | 78.85 | 78.59 | 64.73 | 62.75 | 65.33 |
| LinkNet34 [47] + Orientation [42] | 63.82 | 63.96 | 63.89 | 46.94 | 62.45 | 60.76 | 81.24 | 81.73 | 81.48 | 68.75 | 64.71 | 68.71 |
| MAN [46] | 49.84 | 50.16 | 50.01 | 33.34 | 52.86 | 46.44 | 57.59 | 56.96 | 57.28 | 40.13 | 46.88 | 47.15 |
| RoadTracer [48] | 62.82 | 63.09 | 62.95 | 45.94 | 62.34 | 58.41 | 82.85 | 83.73 | 83.29 | 71.36 | **67.61** | 69.65 |
| OrientationRefine [42] | **64.65** | **64.77** | **64.71** | **47.83** | **63.75** | **63.65** | **83.79** | **84.14** | **83.97** | **72.37** | 67.21 | **73.12** |

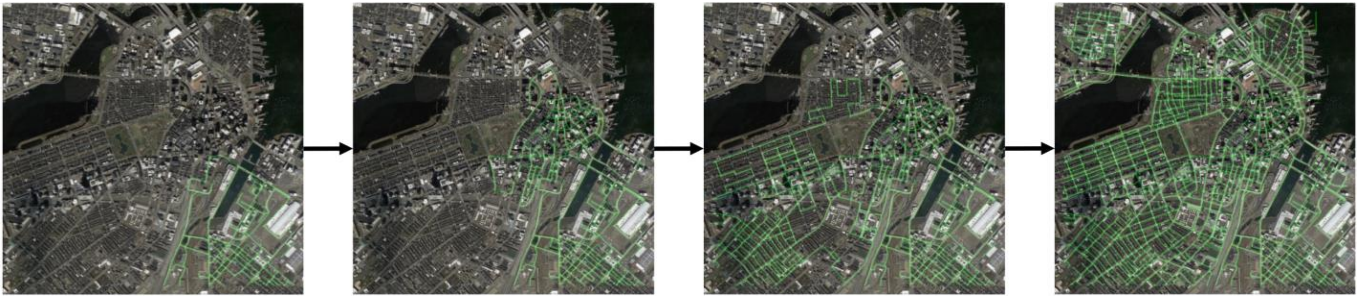

Fig. 8. Rood network extraction on aerial image using iterative graph growing method. The green lines are extracted road [48].

Bastani et al. noticed the same limitation from the DeepRoadMapper. The heuristics perform poorly when there is uncertainty in the road segmentation, which can be caused by occlusion and complex topology such as parallel roads [48]. The CNNs based road segmentation performs poorly as the occlusion area increases, which rises from trees, buildings, and shadows. Prior approaches [36], [55] do not have a solid solution to handle such problems. Bastani et al. proposed a new method, RoadTracer, to address the problems mentioned above and automatically extract the road networks from aerial images [48]. The RoadTracer utilizes an iterative graph construction process, aiming to solve the poor performance caused by occlusion. The RoadTracer has a search algorithm guided by a CNNs based decision function. The search algorithm starts from a known single vertex on the road network and continuously adds vertices and edges to the road network as the search algorithm explores. The CNN-based decision function decides if a vertex or an edge should be added to the road network. In this way, the road graph is generated vertex-by-vertex through an iterative graph growing method. The iterative graph growing method can be visualized in Fig. 8. The RoadTracer approach is evaluated on fifteen city maps, and the results are compared with DeepRoadMapper and another segmentation method Bastani et al. implemented. The RoadTracer can generate a better map network result than the state-of-the-art approach, DeepRoadMapper.

One downside of the iterative graph construction process is the efficiency of generating large-scale road networks. Since this process created the road graph vertex-by-vertex, it will become time-consuming as the road network scale grows. To the best of the author's knowledge, the RoadTracer is the first work that uses the iterative graph growing method to generate road networks. Thus, further research on this method can increase the road network generation efficiency for large-scale road maps.

The evaluation and comparison results of DeepRoadMapper [36], RoadTracer [48], OrientationRefine [42], and other state-of-the-art methods on SpaceNet [43] and DeepGlobe [44] dataset are shown in Table III.

*c) Graph-Generation Method*

Graph-generation methods directly predict the road network graphs from aerial images. This method encodes the input aerial images into vector fields for prediction by neural networks. The prediction is then decoded by a decoding algorithm into graphs. This method has been used to predict road network graphs,

TABLE IV
COMPARISON OF THE THREE TYPES OF ROAD NETWORK EXTRACTION METHODS ON AERIAL IMAGERY USING TOPO-BOUNDARY CITY-SCALE DATASET [62], WITH THE RELAX RATIO OF 2.0, 5.0, AND 10.0 [60]. APLS AND TLTS [63] REFERS TO AVERAGE PATH LENGTH SIMILARITY AND TOO LONG/TOO SHORT SIMILARITY RESPECTIVELY.

| Methods | Precision | | | Recall | | | F1 | | | APLS | TLTS |
|---|---|---|---|---|---|---|---|---|---|---|---|
| | 2.0 | 5.0 | 10.0 | 2.0 | 5.0 | 10.0 | 2.0 | 5.0 | 10.0 | | |
| OrientationRefine | **0.517** | **0.816** | **0.868** | 0.352 | 0.551 | 0.589 | 0.408 | 0.637 | 0.678 | 0.235 | 0.219 |
| Enhanced-iCurb | 0.412 | 0.695 | 0.785 | **0.412** | **0.671** | **0.749** | **0.410** | **0.678** | 0.760 | 0.299 | 0.279 |
| Sat2Graph | 0.460 | 0.484 | 0.604 | 0.128 | 0.240 | 0.293 | 0.159 | 0.304 | 0.374 | 0.037 | 0.030 |
| csBoundary | 0.309 | 0.659 | 0.830 | 0.291 | 0.600 | 0.738 | 0.297 | 0.652 | **0.772** | **0.376** | **0.343** |

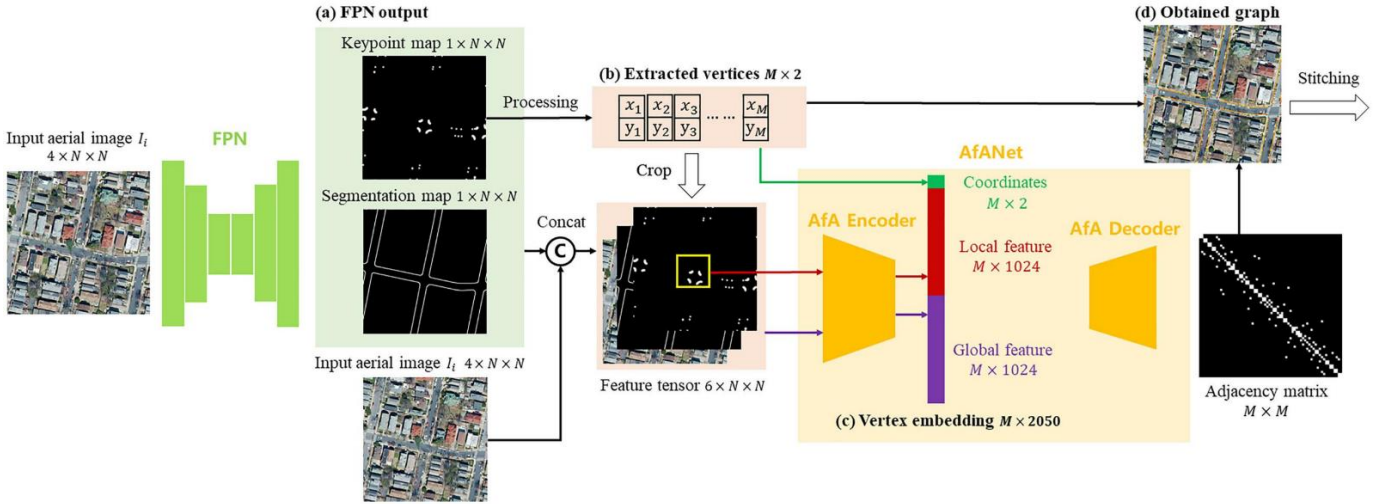

Fig. 9. csBoundary system architecture [60]

including line segments [56], line-shaped objects [57], and polygon-shaped buildings [58].

On top of the graph-generation method, Xu et al. combined the graph-generation method with the transformer [59] and proposed a novel system named csBoundary to automatically extract road boundaries for HD map annotation [60]. The csBoundary system first takes a 4-channel aerial image as the input. It processes the image through Feature Pyramid Network (FPN) [61] to predict the keypoint map and the segmentation map of the road boundaries. From the keypoint map, a set of vertex coordinates are extracted with a length of $M$. The keypoint map, segmentation map, and the input aerial image are combined to form a 6-channel feature tensor. For each extracted vertex, a size of $L \times L$ region of interest (ROI) is cropped and placed at the center of the keypoint map. Xu et al. also proposed the attention for adjacency net (AfANet) [60]. The AfA encoder takes the ROI to calculate the local and global feature vectors, which will be processed by the AfA decoder to predict the adjacency matrix of the extracted vertex for generating the road boundary graph. All obtained graphs will be used to stitch into the final city-scale road boundary graph. The structure of csBoundary can be viewed in Fig. 9.

The segmentation-based method can automatically extract large-scale road networks from aerial images in a very short period of time using CNN. However, the performance of this method heavily depends on the quality of the aerial image. If there is occlusion on the road, which can be caused by shadows or large builds, the segmentation performance will drop. Even with the $A^*$ path planning algorithm assisted in DeepRoadMapper, this method still cannot guarantee a high-performance road network extraction since the shortest path is not always the actual path in real life. The iterative graph growing method, on the other hand, utilizes a search algorithm powered by a CNN-based decision function to enhance the performance of extracting road that has occlusions. However, the iterative graph growing method takes a long time to extract the whole road network since the method builds the road network vertex by vertex. The extraction time of this method will also increase as the size of the road map increases. Since this method extracts the road network in an iterative manner, it also suffers from drifting issues due to the accumulated errors, which makes it challenging for this method to extract large-scale road networks. Graph-generation methods for extracting road networks are still limited to certain shapes of the object since they are heavily relying on the decoding algorithm, which limits their generalization ability. More decoding algorithms are needed to be developed to expand the graph-generation method extraction categories. A performance comparison of the three types of the state-of-the-art methods evaluated on the Topo-Boundary [62] dataset is shown in Table IV using APLS [63] measurement, including OrientationRefine [42] (segmentation-based), Enhanced-iCurb [62] (iterative-graph-growing), Sat2Graph [57] (graph generation), and csBoundary [60] (graph generation).

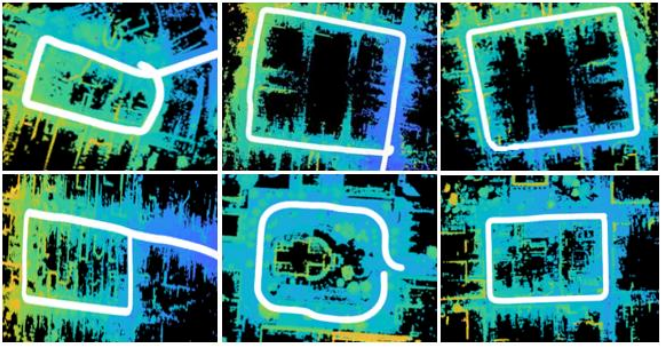

Fig. 10. 3D points clouds are collected in a loop manner. The road point clouds are highlighted in loops with other objects' point clouds in the background [64]

TABLE V
SPECIFIC SCENE HD MAP ARCHITECTURE

| Layer type | Vehicle Terminal | Cloud | Features |
|---|---|---|---|
| Positioning layer | Yes | No | Point cloud and images |
| Road vector and semantic layer | Yes | Yes | Road/lane topology, Road/lane geometry |
| Dynamic object layer | No | Yes | Pedestrians, obstacles, vehicles |
| Real-time traffic Layer | No | Yes | Vehicle speed and position, signal lights status |

*2) Road Extraction on 3D Point Clouds*

Road or lane extraction on 3D point clouds has been widely used in the process of generating HD maps. Lidar point clouds have high precision, usually at millimeter-level accuracy, and contain geometric information of scanned objects. Road extractions using 3D point clouds are done using segmentation.

Ibrahim et al. point out that the 2D road network does not provide any depth clues to the relative positions of objects, and minor infrastructure changes are also not up-to-date in the 2D road network [64]. Instead of building the road network on aerial images, Ibrahim et al. present an HD LiDAR map of the Central Business District (CBD) of Perth, Australia [64]. 3D point cloud data are collected in their work by placing an Ouster LiDAR on top of an SUV and driving the SUV through CBD. The point cloud data is collected in a closed-loop fashion [64] to avoid drifting issues caused by accumulated registration errors, as shown in Fig. 10. A loop detection algorithm is used to extract the point clouds that form the loop, where only the frames that belong to a particular loop will be extracted. The extracted loop point clouds are then pre-processed, including downsampling [65], segmenting ground points [66], and removing the ego vehicle and nearby irrelevant points. The pre-processed loop point clouds are registered and merged using 3D Normal distribution Transformation (NDT) [24]. Post-processing, including spatial subsampling, noise removal, and duplicate points removal and smoothing, is applied to the merged raw point clouds to produce the final extracted road.

Another way of generating a 3D map is proposed by Ding et al. to create HD maps for a specific scene [67]. Their proposal defines a specific scene as a safe and operational environment for autonomous driving applications. In this paper, a part of the university campus is taken as the specific scene for constructing the 3D HD map. Ding et al. divide their HD map architecture into four different layers, including a positioning layer, a road vector and semantic layer, a dynamic object layer, and a real-time traffic layer, see Table V. The positioning layer stores point clouds and images for localization purposes. The road vector and semantic layer stores the road direction of travel, road type and on-road object. In this layer, the OpenDRIVE file format is used. The dynamic object layer, as the name implies, stores highly dynamic perceptual information about objects such as pedestrians, obstacles, and vehicles. This layer is updated at a higher frequency to provide feedback from the surrounding environment. The real-time traffic layer stores the real-time traffic data such as vehicle speed and position and traffic signal lights status. The 3D HD map is created using the NDT algorithm with a digital 3D scene of the actual scene as the reference. The map results are shown in Fig. 11. More details about their mapping procedures can be found in [67].

*3) Road/Boundary Extraction with Sensor Fusion Methods*

Road extraction on 2D aerial images and 3D point clouds both have limitations. Due to poor lighting conditions, occlusions caused by the roadside furniture, and various terrain factors, road networks extracted from satellite and aerial images are usually inaccurate and incomplete. Feature extractions on 3D point clouds also face occlusion and point density variation issues, which lead to inaccurate and incomplete road extraction. Limitations of employing a single data source are noticeable in extracting roads or road boundaries. Thus, researchers have been employing multi-source data to extract and complete roads or road boundaries. Gu et al. [68] employed lidar's imageries and camera-perspective maps and constructed a mapping layer to transfer the features of lidar's imageries view to the camera's perspective imagery view. This approach enhanced the road extraction performance in the camera's perspective view. Gu et al. [69] also proposed a conditional random forest (CRF) framework to fuse lidar point clouds and camera images to extract both range and color information of road networks. In [70], a fully-convolutional network (FCN) is designed to merge feature maps learned from lidar-camera data based on the residual fusion strategy for road detection. Li et al. [71] took a different approach to build road maps by fusing GPS trajectories and remote sensing images. This approach used a transfer learning-based neural network to extract road features from images and used U-Net to extract the road centerline. Additionally, a tightly-coupled perception-planning framework was designed in [72] to detect road boundaries by using GPS-camera-lidar sensor fusion.

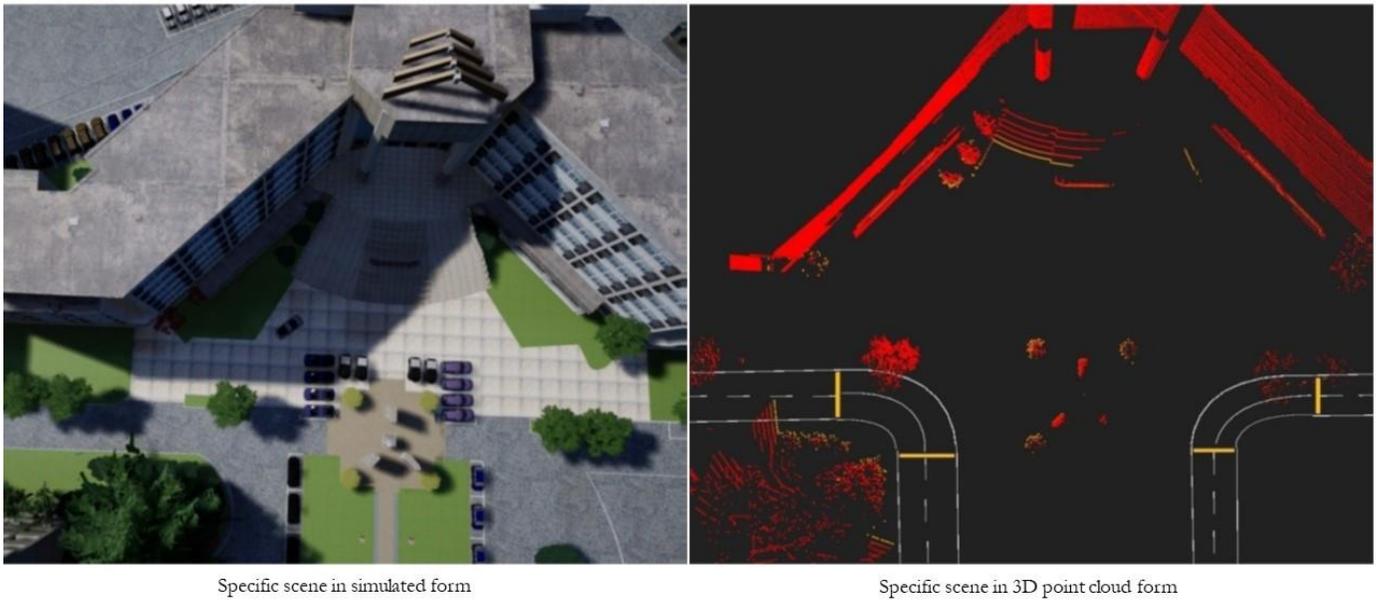

Fig. 11. Digital 3D scene of campus building (left) and its HD map (right) [67]

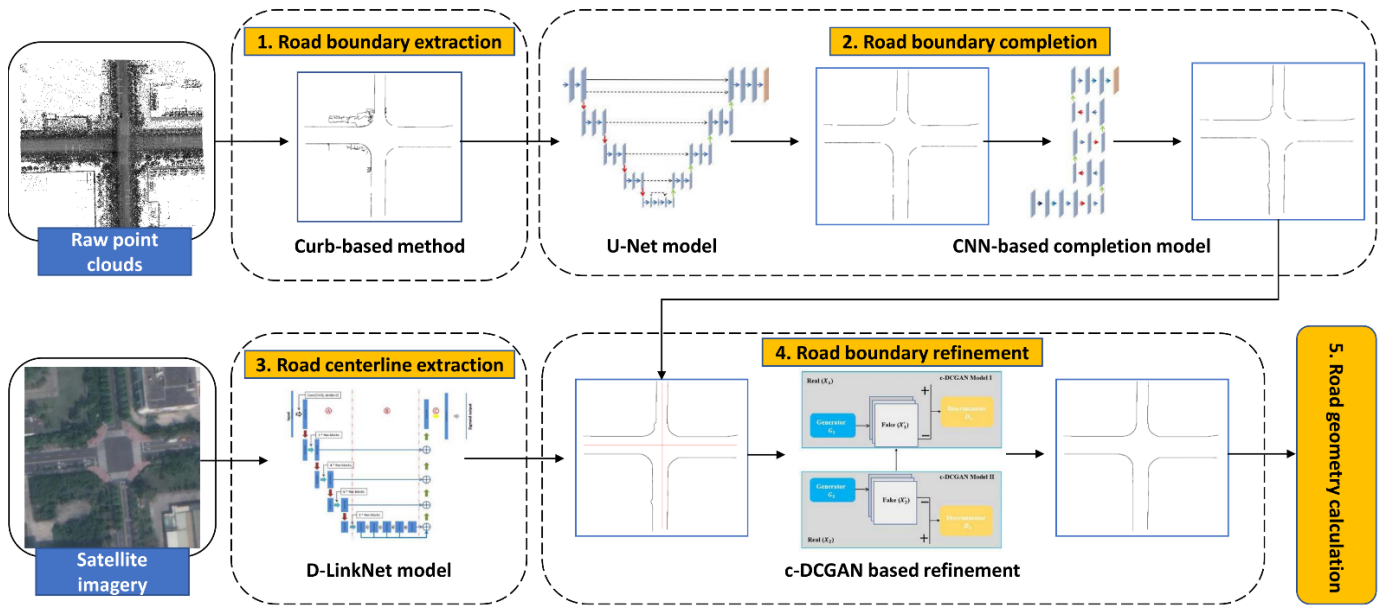

Fig. 12. The architecture of BoundaryNet [73]: 1. The road boundary extraction from the raw point clouds using the curb-based method. 2. Road boundary completion by applying a U-shaped encoder-decoder model and a CNN-based completion model. 3. Road centerline extraction from satellite imagery by using D-LinkNet model [76]. 4. Road boundary refinement by using a c-DCGAN model. 5. Road geometry calculation based on extracted road boundary.

Ma et al. also proposed a novel deep learning framework, named BoundaryNet, to extract road boundaries and fill the existing gap in road boundary data caused by occlusion using both laser scanning point clouds and satellite imagery [73]. This method used a curb-based extraction method to extract the road boundaries and employed a modified U-net [74] model to remove the noise point clouds from the road boundary point clouds. A CNN-based road boundary completion model was then applied to the extracted road boundaries to fill some of the gaps. Inspired by an image-to-image translation method, Generative Adversarial Networks (GAN) [75], a conditional deep convolutional generative adversarial network (c-DCGAN) was designed to extract more accurate and complete road boundaries, with the assistance of road centerlines extracted from the satellite images. The architecture of the proposed method can be viewed in Fig. 12, with D-LinkNet model for road centerline extraction [76].

TABLE VI.
ROAD NETWORK EXTRACTION METHODS COMPARISON

| Comparison | Method/Data Source | | |
|---|---|---|---|
| | *Extraction on aerial images* | *Extraction on 3D point clouds* | *Extraction with Sensor fusion* |
| Advantages | Large road network coverage, Time-saving in data collection | Rich geometric information, High accuracy (millimetre-level), Able to capture various kinds of objects | Rich geometric information, Multi-sensors solve data imperfection issues, Higher accuracy than a single data source |
| Disadvantages | Lack of depth/elevation information, Performance heavily depends on the aerial image quality | Time-consuming in data collection, 3D LiDAR is costly, Performance depends on the point clouds quality | Multi-sensors system is costly, Time-consuming in data collection |

*4) Other Methods*

There are also different methods to extract road networks. Schreiber et al. and Jang et al. [77], [78] took a different approach by extracting the road from camera images instead of aerial images. Former did 3D reconstruction from camera images, and the latter designed a fully convolutional network (FCN) to detect and classify the road. Both approaches can be applied to small-scale HD maps but not large or city scale HD maps due to tremendous labor work and time consumption from data collection. [79] lists more machine learning-based road/lane extraction methods and Aldibaja et al. [80] also proposed a 3D point cloud accumulation method, which is also worth learning but will not be discussed in detail in this review.

Road extraction can be done through different data sources, including camera images, satellite and aerial images, lidar point clouds, and GPS trajectories. Satellite and aerial images can cover large-scale maps, making road extraction for city-scale road networks highly efficient. However, road networks extracted from satellite and aerial images do not contain depth or elevation information. The performance of road extraction from aerial images also heavily depends on the quality of the images. Factors such as poor lighting conditions, occlusions caused by roadside furniture, and various terrain factors can reduce the performance of the extraction. Road extraction from 3D point clouds, in contrast, has more geometric information and high accuracy level (millimetre-level), but it also faces the occlusion issue, which leads to incomplete road extraction. Point density variation issues also lead to inaccurate road extraction. Sensor fusion methods have then been introduced to further enhance road extraction performance by fusing different data sources such as aerial images, GPS data, camera images, and lidar point clouds. Sensor fusion methods have outperformed methods using a sole data source and achieved remarkable results in road extraction. The comparison of three methods is summarized in Table VI.

*B. Road Markings Extraction*

Road markings/pavement markings are signs on concrete and asphalt road surfaces [81]. They are usually painted with highly retro-reflective materials, making them noticeable to human vision and sensors of autonomous vehicles. Road markings are essential features on HD maps to provide the ego vehicle with information about the direction of traffic, turning lanes, travelable and non-travelable lanes, pedestrian crossings, etc. [82]. Similar to road extraction methods, road marking extraction can also be done using 2D images or 3D point clouds.

*1) Road Marking Extraction on 2D Images*

Traditionally, roading marking extractions on 2D images are achieved through image processing and computer vision. Images containing road markings are first denoised and enhanced to make the road markings clear and obvious and highlight the contrast between the target and the background area. Then, the target road marking is extracted using image processing and computer vision methods such as edge-based detection (such as Roberts, Sobel, Prewitt, Log, and Canny), threshold segmentation (such as Otsu's method and iterative method), k-means clustering, and regional growth method [83]. The traditional methods have achieved remarkable performance in extracting road markings from pavement or concrete roads. However, simple extractions without proper recognition of different road markings are not efficient enough for the ego vehicle to understand the road rules. With the introduction and rapid development of CNNs, methods involving CNNs have been widely developed and employed in detecting and recognizing road markings. Road marking extraction and recognition on 2D images is usually conducted in two different approaches. One is to exploit the front-view images captured by vehicle-mounted cameras. The other is to extract road markings from aerial images. An example of both of them is shown in Fig. 13.

low-level features of objects (such as colors, shapes, boundaries and so on) and filter out the interference information which is likely to affect the model learning process. The second

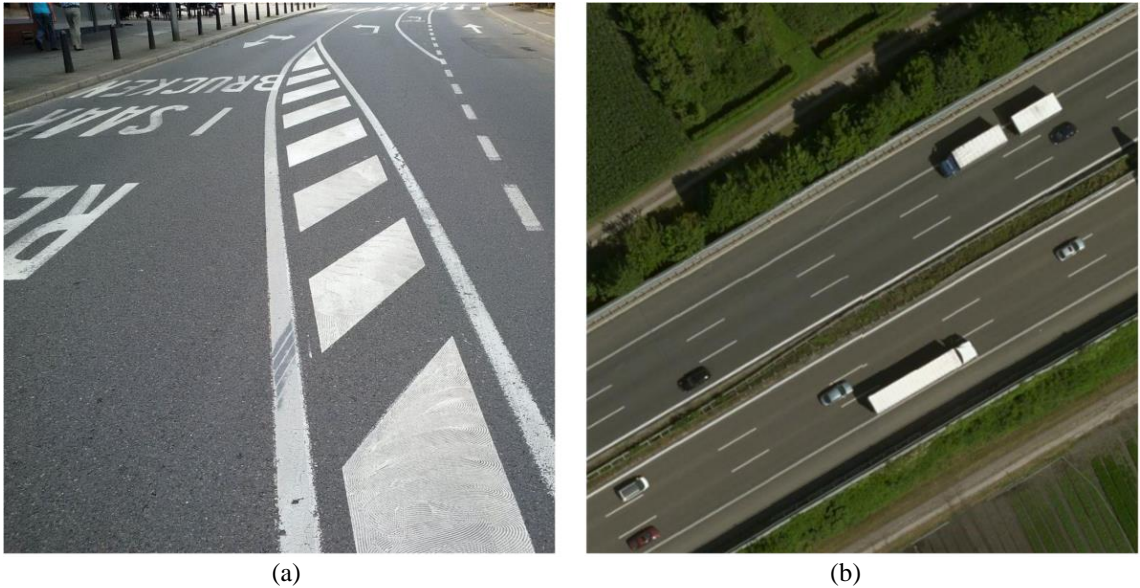

Fig. 13. Road markings on the front-view image VS road markings on the aerial image

*a) Road Marking Extraction on Front-view Images*

The front-view images have been intensively used for road marking extraction due to their cost-effectiveness and convenience. Several methods have been proposed to detect the lane line markings. Zhang et al. [84] proposed Ripple Lane Line Detection Network (RiLLD-Net) to detect common lane line markings and Ripple-GAN to detect complex or occluded lane line markings. RiLLD-Net is a combination of U-Net [74], the residual modules with the skip connection, and the quick connections between encoders and decoders. An original image containing lane line markings is pre-processed into a gradient map using the Sobel edge detection filter [85]. Both the original image and the gradient map are fed into the RiLLD-Net to remove the redundant interference information and highlight the lane line markings. The proposed Ripple-GAN is a combination of Wasserstein GAN (WGAN) [86] and RiLLD-Net. An original lane line marking image added with white Gaussian noise is sent to WGAN to produce the segmented lane line marking results. The segmentation results together with the gradient map are sent to RiLLD-Net to further enhance the lane line marking detection result. Furthermore, [87] proposed a spatio-temporal network with double Convolutional Gated Recurrent Units (ConvGRUs) [88], [89] for lane line detection. Instead of taking one image at a time, the network takes multiple captures consisting of lane line markings at continuous time stamps as inputs. Each of the two ConvGRUs has its own function. The first ConvGRU, also called Front ConvGRU (FCGRU), is placed at the encoder phase and used to learn the ConvGRU, also called middle ConvGRU, contains multiple ConvGRUs. It is placed between the encoder and decoder phases and used to thoroughly learn the spatial and temporal driving information of continuous driving images that FCGRU produces. The network then concatenates the downsampling layers from the encoder and the upsampling layers from the decoder to produce the final lane line marking detection. Additionally, other methods have also been proposed to solve the lane line detection and extraction problem, such as graph-embedded lane detection [90], progressive probabilistic hough transform based lane tracking [91], SALMNet [92], segmentation based lane detection [93], and mask R-CNN instance segmentation model [94]. Table VII summarizes the evaluation results of methods [84], [87], [90], [92], [93] on the TuSimple dataset [95] for clear comparisons. Previous state-of-the-art methods, including SCNN [96], LaneNet [97], and Line-CNN [98] are also included in Table VII to show the improvement of the current state-of-the-art methods. The bold highlighted values in the Table VII represent the best results. Based on the comparison, Ripple-GAN is currently the best state-of-the-art method.

*b) Road Marking Extraction on Aerial Images*

Satellite and aerial images can be used for not only road network extraction but also road marking extraction. Azimi et al. [99] proposed Aerial LaneNet to directly extract road markings from aerial images. The proposed network contains a symmetric fully convolutional neural network (FCNN). The original aerial images are chopped into multiple patches before sending to Aerial LaneNet. Aerial LaneNet predicts a semantic segmentation of every input patch and produces a binary image for each patch that denotes which pixel is from lane markings and which one is from the background. All binary images/patches are stitched together to build the final road marking image which has the same resolution as the input

### 2) Road Marking Extraction on 3D Point Clouds

Road markings extraction on 3D point clouds is usually done in two different methods, bottom-up method and top-down method [81]. Bottom-up methods directly extract the road marking by differentiating the road marking point cloud from image. The model also utilizes the discrete wavelet transform (DWT) to achieve multiscale and full-spectrum domain analysis. Similarly, Kurz et al. [100] designed a wavelet-enhanced FCNN to segment Multiview high-resolution aerial imagery. The imagery2D segments are further used to create a 3D reconstruction of road markings based on the least-squares line-fitting.

TABLE VII.
EVALUATION RESULTS ON TUSIMPLE DATASET OF METHODS FOR ROAD MARKING EXTRACTION ON FRONT-VIEW IMAGES

| Methods | Accuracy | FP | FN | Precision | Recall | F1 |
|---|---|---|---|---|---|---|
| SCNN [96] | 0.9653 | 0.0617 | **0.018** | 0.7252 | 0.9653 | 0.8282 |
| LaneNet [97] | 0.9638 | 0.078 | 0.0244 | 0.7554 | 0.9638 | 0.8470 |
| Line-CNN [98] | 0.9687 | 0.0442 | 0.0197 | N/A | N/A | N/A |
| Spatio-Tempeoral [87] | **0.9804** | N/A | N/A | 0.8750 | 0.9531 | 0.9124 |
| SALMNet [92] | 0.9691 | 0.0263 | 0.0252 | N/A | N/A | N/A |
| PINet [93] | 0.9675 | 0.0310 | 0.0250 | N/A | N/A | N/A |
| Graph-Embedded [90] | 0.9637 | N/A | N/A | N/A | N/A | N/A |
| Ripple-GAN [84] | N/A | **0.0048** | 0.0289 | **0.9806** | **0.9728** | **0.9767** |

Yu et al. also proposed a self-attention-guided capsule network, called MarkCapsNet [101], to extract road markings from aerial images. The proposed network, combining the capsule formulation and the HRNet [102] architecture, can extract feature semantics at different scales by involving three parallel branches with different resolutions. The capsule-based self-attention (SA) module is also designed and integrated into each branch of the MarkCapsNet to further enhance the representation quality of the generated feature map for road marking extraction. Additionally, Yu et al. also created a large-scale aerial image dataset for road marking extraction applications called AerialLanes18, which can be used as the benchmark for testing different methods for road marking extraction in the future. MarkCapsNet and other road marking extraction models, [99], [102]–[105], were experimented with and compared on two datasets, UAVMark20 and AerialLanes18. The results, in Table VIII, show that MarkCapsNet has achieved state-of-the-art performance.

Road marking extraction on the front-view image method has a much smaller field of view compared to aerial images, and the detection/processing time is also longer than using the existing extracted road markings on the aerial images. However, it is flexible to changes in road markings such as wearing and occlusion since the detection is based on real-time camera images. In contrast, road marking extraction on aerial images can extract road markings for larger scales, and it stores the extracted road markings in the HD map to reduce the detection time. However, it is sensitive to the data imperfection caused by lighting conditions, occlusions, and road marking wear.

the background point cloud. In contrast, top-down methods detect the pre-defined geometric models using CNN and reconstruct the road markings based on the detection.

#### a) Bottom-up Method

The bottom-up methods use deep learning algorithms to directly extract the road markings from the raw 3D point clouds based on object detection and segmentation. The thresholding-related method and its extensions, including multi-thresholding and multi-thresholding combined with geometric feature filtering, are widely used for road marking extraction [106]–[109]. [103], [110] convert 3D point clouds into 2D georeferenced intensity images before extracting road markings to dramatically reduce the computational complexity but lose the pose (position and orientation) or spatial information. To fill the gap of missing pose or spatial information from extracted road markings, Ma et al. proposed a capsule-based network for road marking extraction and classification [111]. The proposed approach processes the data by filtering out the off-ground feature point clouds such as poles, traffic lights, and trees to reduce the computational complexity. The processed 3D point clouds are converted to 2D georeferenced intensity raster images using the inverse distance weighting (IDW) algorithm. Inspired by the capsule network proposed in [112], Ma et al. proposed a U-shaped capsule-based network that can learn not only the intensity variance from the raster images but also the pose and the shape of road markings. A hybrid capsule network is proposed to categorize different road markings into different classes. The performance of the proposed approach for road marking extraction in urban, highway, and underground garages has achieved the best state-of-the-art result, with an F1 score of 92.43% and a precision score of 94.11%. The performance comparison of [103], [107], [111], [113] on a customized road marking dataset can be viewed in Table IX, where the bold highlighted values are the leading results.

TABLE VIII.
ROAD MARKING EXTRACTION RESULTS OF DIFFERENT MODELS

| Model | Dataset | Precision | Recall | IoU | F1-score |
|---|---|---|---|---|---|
| MarkCap-sNet [101] | UAVMark20 | **0.7412** | **0.7622** | **0.6020** | **0.7516** |
| | AerialLanes18 | **0.7663** | **0.7851** | **0.6334** | **0.7756** |
| U-Net [103] | UAVMark20 | 0.6401 | 0.6617 | 0.4823 | 0.6507 |
| | AerialLanes18 | 0.6752 | 0.6836 | 0.5144 | 0.6794 |
| Aerial LaneNet [99] | UAVMark20 | 0.6625 | 0.6811 | 0.5057 | 0.6717 |
| | AerialLanes18 | 0.6973 | 0.7155 | 0.5460 | 0.7063 |
| HRNet [102] | UAVMark20 | 0.7112 | 0.7295 | 0.5628 | 0.7202 |
| | AerialLanes18 | 0.7378 | 0.7542 | 0.5948 | 0.7459 |
| SegCaps [104] | UAVMark20 | 0.6672 | 0.6773 | 0.5063 | 0.6722 |
| | AerialLanes18 | 0.7012 | 0.7113 | 0.5459 | 0.7062 |
| SA-CapsFPN [105] | UAVMark20 | 0.7158 | 0.7347 | 0.5688 | 0.7251 |
| | AerialLanes18 | 0.7416 | 0.7591 | 0.6003 | 0.7502 |

*b) Top-down Method*

The top-down method uses the existing object detection algorithm to detect and locate the road marking geometric models. It reconstructs the road markings on the 3D point clouds based on the detection and localization. Prochazka et al. [114] used the spanning tree usage method to automatically extract lane markings from point clouds into a polygon map layer. The proposed approach makes the ground point detection [115] on the raw point clouds and identifies the detection using the spanning tree. The lane marking is then reconstructed into the vector form after the detection and identification. This method can detect lane markings but not other kinds of road markings, such as road direction markings and pedestrian crossing markings.

Mi et al. proposed a two-stage approach for road marking extraction and modeling using MLS point clouds [81]. Their approach utilizes YOLOv3 [116] algorithm to detect the road markings and provide the semantic labels on each detection. During the reconstruction and modeling process, an energy function (1) is proposed to determine the fine pose and the scales of the road markings in the raw 3D point clouds.

$$E_{fine}(x, y, z, \varphi, \theta, s) = \alpha * \frac{1}{N} \sum_{i=1}^{N} I_i + (1-\alpha) * \frac{1}{N_e} \sum_{j=1}^{N_e} G_j \quad (1)$$

$$G = \frac{1}{N_{in}} \sum_{i=1}^{N_{in}} I_i - \frac{1}{N_{ex}} \sum_{i=1}^{N_{ex}} I_i \quad (2)$$

where $(x, y, z, \varphi, \theta, s)$ is the location and the orientation of the road marking template, $\alpha$ is the weight, $I_i$ is the intensity of the $i$-th point in the road marking template, $N$ is the number of points in the template, $N_e$ is the number of the sample points on the board of the template, and $G_j$ is the intensity gradient of the $j$-th border point, which is defined by (2), where $N_{in}$ and $N_{ex}$ are the internal and external number of points, respectively, within a searching radius [81]. The road markings can be segmented and modeled precisely by maximizing the energy function. The road marking extraction performance is further improved with the candidate re-ranking strategy that combines the detection confidence and the localization score to produce the ultimate road marking models on the 3D point clouds. The candidate re-ranking strategy aims to provide an optimal selection when road markings overlap exists or a mirror-symmetric twin exists, as shown in Fig. 14. This approach can precisely extract twelve types of road markings specified in the Chinese national standards for road traffic signs and markings.

In summary, the bottom-up method can directly extract the road markings from MLS point clouds and speed up the road marking extraction progress, but it is sensitive to imperfect raw data. In contrast, imperfect raw data has less effect on the top-down method, but the top-down method is time-consuming since the detection leads to huge search space. To the author's best knowledge, there are not a lot of approaches for road marking extraction utilizing the top-down method compared to the bottom-up method. Thus, road marking extraction with the top-down method requires further research.

*C. Pole-like Objects Extraction*

TABLE IX.
ROAD MARKING EXTRACTION RESULTS COMPARISON USING DIFFERENT BOTTOM-UP METHODS [124]

| Method | Road scene | Precision | Recall | F1-score |
|---|---|---|---|---|
| Ma et al. [107] | Urban | 62.63 | 53.19 | 57.53 |
| | Highway | 70.13 | 65.54 | 67.76 |
| | Underground garage | 68.73 | 59.42 | 63.74 |
| | Average | 67.16 | 59.38 | 63.01 |
| Wen et al. [103] | Urban | 92.15 | 89.33 | 90.72 |
| | Highway | 95.97 | 87.52 | 91.55 |
| | Underground garage | **91.95** | 90.07 | 91.00 |
| | Average | 93.36 | 88.97 | 91.09 |
| Cheng et al. [113] | Urban | 27.35 | 33.82 | 30.24 |
| | Highway | 30.57 | 34.10 | 32.24 |
| | Underground garage | 24.52 | 29.03 | 26.59 |
| | Average | 27.48 | 32.32 | 29.69 |
| Ma et al. [111] | Urban | **94.92** | **90.25** | **92.52** |
| | Highway | **96.14** | **91.13** | **93.57** |
| | Underground garage | 91.26 | **90.17** | 91.20 |
| | Average | **94.11** | **90.52** | **92.43** |

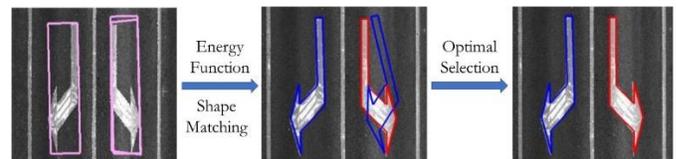

Fig. 14. Road marking extraction using energy function and candidate re-ranking strategy [81]

In an HD map, pole-like objects such as traffic lights, traffic signs, streetlights, trees, and telephone line poles are essential for the road environment. They can help with localization (different shapes from other road furniture) and motion planning (traffic light signals provide traffic flow conditions). Pole-like object extractions are usually done through segmentation and classification on MLS 3D point clouds.

Various approaches for pole-like object extractions have been developed in previous years. Lehtomäki et al. [117] proposed to detect vertical pole-like objects using MLS 3D point clouds with combined segmentation, clustering, and classification methods. El-Halawany et al. [118] adopted a covariance-based procedure to perform pole-like object segmentation on terrestrial laser scanning point clouds and find their dimensions. Yokoyama et al. [119] applied Laplacian smoothing using the k-nearest neighbors graph and used principal component analysis to recognize points on pole-like objects with various radii and tilt angles. Pu et al. [120] presented a framework for structure recognition from MLS 3D point clouds and extracted pole-like objects based on the characteristics of point cloud segments like size, shape, orientation and topological relationships. Cabo et al. [121] proposed a method to detect pole-like objects by spatially discretizing the point cloud with regular voxelization and analyzing and segmenting the horizontally voxelized point clouds. Ordóñez et al. [122] added a classification module on top of [121] to distinguish between the different types of poles in the point cloud. Yu et al. [123] proposed a semi-automated extraction of streetlight poles by segmenting MLS 3D point clouds into the road and non-road surface points and extracting the streetlight poles from non-road segments using a pairwise-3D shape context. Zheng et al. [124] proposed a new graph-cut-based segmentation method to extract the streetlight poles from MLS 3D point clouds followed by a Gaussian-mixture-model-based method for recognizing the streetlight poles.

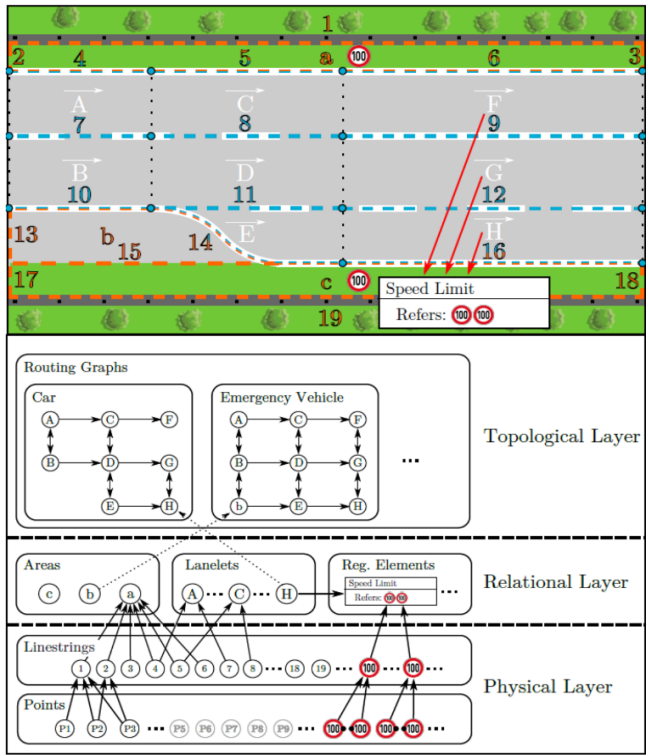

Fig. 15. Lanelet2 Map Structure [131]: physical layer defines physical elements formed by points and linestrings, such as pole-like objects, markings, and boundaries. The Relational layer defines areas, lanelets, and regulatory elements, such as buildings, motorways, the direction of travel, and traffic rules. The topological layer defines the topological relationships between elements from the first two layers. The topological layer in the figure shows the routing for normal vehicles and emergency vehicles.

Plachetka et al. [125] recently proposed a deep neural network (DNN) based method that recognizes (detects and classifies) pole-like objects in the LiDAR point clouds. Inspired by [126], the proposed DNN architecture consists of three stages: encoder, backbone, and classification and regression head. Raw 3D point clouds are pre-processed into cell feature vectors which are the inputs of the encoder stage. Instead of having only one encoder stage in [127], Plachetka et al. added another encoder stage following [128], to improve the representational power of the cell feature vectors. The encoder encodes the cell feature vectors into a spatial grid as the input to the backbone. The backbone stage adopts and modifies the feature pyramid architecture [61] to include more context, improving the model performance in detecting small objects [129]. The backbone stage concatenates low-level features from the downstream path and high-level features from the upstream path to further enhance the representational power of the input grid. The downstream path utilizes the convolutional layer, and the upstream path uses the transposed convolutional layer. The output feature grid becomes the classification and regression head stage input. The architecture adopts the SSD [130] method to compute prediction in the last stage. The method achieves a mean recall, precision, and classification accuracy of 0.85, 0.85, and 0.93, respectively. Details on the proposed DNN architecture, datasets, and training procedure are out of the scope of this review and are not discussed further. The detection categories include protection poles, traffic sign poles, traffic light poles, billboard poles, lampposts and trees.

In summary, pole-like objects in HD maps are important features for localization due to their special shapes. Pole-like object extractions are mainly done on 3D point clouds, thus the performance of the extraction also depends on the quality of the point clouds. Thus, further research on improving the pole-like object extraction performance on imperfect data is required.

## V. FRAMEWORK FOR HD MAPS

With the increasing complexity of the HD maps and the number of environmental features that require extraction, it is necessary to have good software in the form of a framework to sufficiently store the relevant information in a map and ensure a consistent view of the map [131]. In this section, three popular open-source frameworks for creating HD maps are introduced, including Lanelet2 [131], OpenDRIVE [132], and Apollo Maps [133].

## A. Lanelet2

Lanelet2 is an extension and generalization of Liblanelet, also known as Lanelet, developed for the Bertha Drive Project. Lanelet2 map adopts the exiting format from Lanelet, the XML-based (extensible markup language) OSM (Open Street Map [2]) data format. Open Street Map is a free online map editing tool that is constantly updated and contributed by map editors worldwide. However, the actual data format of a map is considered to be irrelevant and interchangeable as long as this format can be converted to Lanelet2 format without any information loss.

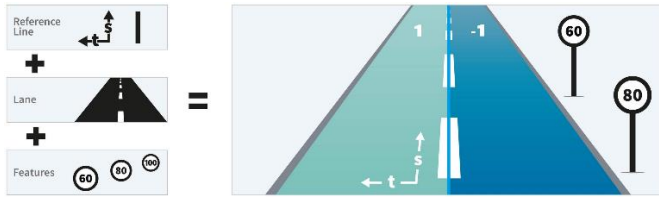

Fig. 16. ASAM OpenDRIVE HD map structure [110]

A Lanelet2 map contains three layers: the physical, the relational, and the topological layer, as shown in Fig. 15. The characteristics of these three layers are similar to the ones defined by HERE. The first physical layer consists of two elements, points and line strings. Points are the basic element of the map. It can represent a vertical structure such as poles as single points, a lane, or an area as a group of points. Line strings are constructed as an ordered array of two or more points, where linear interpolation is used between every two points. The physical layer, as the name implies, defines the detectable elements such as traffic lights, markings, curbstone, etc. The second relational layer consists of three elements, lanelets, areas, and regulatory elements. Lanelets define different road types, such as regular lanes, pedestrian crossings, and rails. Lanelets are also associated with traffic rules which do not change within a lanelet. It is defined by exactly one left and one right line string as two borders with opposite directions. The directions in the pair of line strings are interchangeable by changing the left border to the right border (and vice versa). Areas are constructed by one or more line strings to form a closed barrier and usually represent static constructions such as buildings, parking lots, playgrounds, and grass spaces. As the name implies, the regulatory elements define the traffic rules to regulate the ego vehicle. Lanelets and areas can have one or more regulatory elements, such as speed limits and restrictions. Dynamic rules such as turning restrictions based on time of the day can also be added as the regulatory element.

Lanelet2 is a simple and powerful framework to support HD maps. It is also frequently used with Autoware Auto [27] to create vector maps for the HD map. More details about the Lanelet2 framework can be found in [131], [134].

## B. OpenDRIVE

OpenDRIVE is an open-source framework for describing road networks and creating HD maps developed by the Association for Standardization of Automation and Measuring Systems (ASAM). It also uses XML file format to store the map information.

In an ASAM OpenDRIVE map, there are three elements/layers, Reference Line/Road, Lane, and Features, see Fig. 16. Unlike Lanelet2 maps, which use points to describe and construct map features, OpenDRIVE uses geometric primitives including straight lines, spirals, arcs, cubic polynomials, and parametric cubic polynomials to describe the road shape and direction of travel. Those geometric primitives are called Reference Line. The reference line is the key component of every OpenDRIVE road network, as all lanes and features are built along the reference line. The second element, Lanes, attaches to the reference line and represents the drivable path on the map. Every road contains at least one lane with a width greater than 0. The number of lanes on each road depends on the actual traffic lanes and has no limits. A center lane with 0 widths is required when constructing lanes along roads as a reference for lane numbering, see Fig. 17. The center lane defines the driving direction on both sides based on the road types, and it can be either opposite or the same direction. In fig. 17, since there is no offset between the center lane and the reference line, the center lane is coincident with the reference line. The last element, Features, contains objects such as signals and signs that are associated with traffic rules. However, unlike Lanelet2, the dynamic content is not covered by ASAM OpenDRIVE. A well-documented user guide on OpenDRIVE provided by ASAM can also be found in [135].

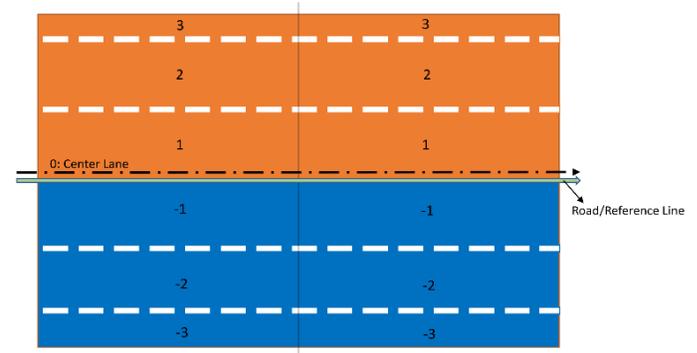

Fig. 17. Center lane for roads with lanes of different directions of travel

## C. Apollo Maps

Apollo Maps is the HD map created by Baidu Apollo, a leading autonomous driving platform in China. Apollo HD maps also use the OpenDRIVE format, but a modified version specifically for Apollo. Apollo simply uses points, unlike OpenDRIVE, using geometric primitives like lines, spirals, and arcs to define roads. Like the points in Lanelet2, each of the points stores latitude and longitude values, and a list of these points defines the road boundaries. In Apollo HD maps, there are typically five different elements: 1. Road elements contain features like road boundaries, lane type, and lane direction of travel; 2. Intersection elements have the intersection boundaries; 3. Traffic signal elements include traffic lights and signs; 4. Logical relationship elements contain the traffic rules; 5. Other elements include crosswalks, street lights, and buildings [133]. To construct the HD maps, Baidu Apollo divides the generation process into five steps: data sourcing, data processing, object detection, manual verification, and map production, as shown in Fig. 18. The detailed Apollo HD map generation procedures can be found in [133].

From the author's point of view, Apollo map is a more advanced and complex version of OpenDRIVE. Apollo maps contain elements that were not originally in OpenDRIVE, such as no parking area and crosswalk. Apollo maps also require more data to define lanes than OpenDRIVE. OpenDRIVE only needs to specify the lane width, while Apollo requires points to describe the lane borders. In order to use OpenDRIVE Maps in Apollo, one can use the method provided here [136] to convert the OpenDRIVE format to Apollo format. Lanelet2 map can also be converted to OpenDRIVE map format. Carla, an open-source simulator for autonomous driving, provides a PythonAPI for converting the OSM map to the OpenDRIVE map [137].

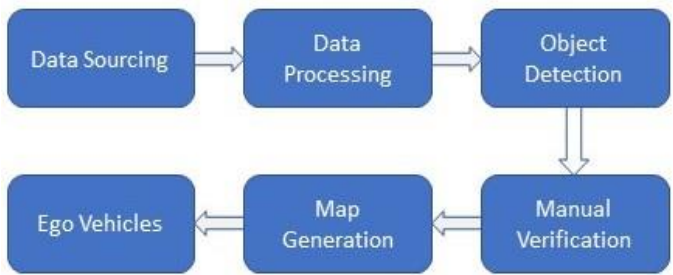

Fig. 18. Apollo Maps Generation Process

## VI. LIMITATIONS AND OPEN PROBLEMS

HD map generation technologies have experienced rapid development in recent years. However, there are still limitations. Feature extractions on 2D images can quickly generate features like lane lines and road marking for large-scale maps using aerial images, but the extraction does not contain altitude or depth information. Altitude or height information can be manually added to the 2D map to create the 2.5D map by matching road network GPS data with the collected GPS data and adding the corresponding altitude. However, it still lacks depth information, which is extremely important when the ego vehicle drives around obstacles. 2D HD maps are also not sensitive to minor changes to infrastructure, which will not keep the map up-to-date. Feature extraction on MLS 3D point clouds is a more common and powerful approach to adding detailed road information to HD maps. HD maps with extracted 3D features provide the depth information and updated environmental information but require expensive LiDAR and high computational costs. It is also time-consuming to collect usable point cloud data. An example can be found in [64] that Ibrahim et al. spent three two-hour sessions of driving an SUV to collect usable point cloud data. [138] used crowdsourcing method to keep HD maps updated, but the crowdsourced method is not always an available solution for individual researchers. It will be challenging to collect point cloud data for city-scale maps.

The limitations lead to open problems that are: 1. Adding more features such as depth information to 2D maps and keeping them updated; 2. Increasing the efficiency of the 3D map generation process and making 3D mapping for large-scale HD maps possible without consuming too much time and computational power. One solution is to integrate the road network and the point cloud for HD map generation, which can be done using Autoware. Moreover, to the best of the author's knowledge, there are not a lot of proposed approaches for feature extractions on sidewalks. This is critical and in demand, as some autonomous systems are either designed to drive on the sidewalk or require feasibility tests on sidewalks before testing on motorways. Furthermore, the completion and finishing of HD maps (merging all modules and features into an HD map) are still commercialized methods developed by mapping companies. This is still an open problem for academic and individual researchers, which requires further research and conclusion

## VII. CONCLUSION

In this review, recent HD map generation technologies for autonomous driving have been analyzed. The basic common structures of HD maps are summarised in three layers: 1. Road Model; 2. Lane Model; 3. Localization Model. Data collection, 3D point cloud generation, and feature extraction methods for generating HD maps, including road networks, road markings, and pole-like objects, are introduced and compared. Limitations of those approaches are also discussed. The frameworks for supporting HD maps are also introduced, including Lanelet2, OpenDRIVE, and Apollo. Some useful tools for converting map formats among three frameworks are also provided.

Some challenging issues for further research and development are 1. Adding more features such as depth information to 2D maps and keeping them constantly updated; 2. Increasing the efficiency of the 3D map generation process and making 3D mapping for large-scale HD maps possible without consuming too much time and computational power. 3. Creating feature extraction methods for sidewalks. 4. Uncommercialized methods for completion and finishing of HD maps.

ACKNOWLEDGMENT

This study was supported by the start-up funding under Dr. Xianke Lin.

## REFERENCES

[1]  J. Ziegler, P. Bender, M. Schreiber, H. Lategahn, T. Strauss, C. Stiller, T. Dang, U. Franke, N. Appenrodt, C. G. Keller, E. Kaus, R. G.

Herrtwich, C. Rabe, D. Pfeiffer, F. Lindner, F. Stein, F. Erbs, M. Enzweiler, C. Knoppel, J. Hipp, M. Haueis, M. Trepte, C. Brenk, A. Tamke, M. Ghanaat, M. Braun, A. Joos, H. Fritz, H. Mock, M. Hein, and E. Zeeb, "Making bertha drive-an autonomous journey on a historic route," *IEEE Intell. Transp. Syst. Mag.*, vol. 6, no. 2, pp. 8–20, Apr. 2014, doi: 10.1109/MITS.2014.2306552.

[2] R. Liu, J. Wang, and B. Zhang, "High definition map for automated driving: overview and analysis," *J. Navig.*, vol. 73, no. 2, pp. 324–341, Mar. 2020, doi: 10.1017/S0373463319000638.

[3] S. Ulbrich, A. Reschka, J. Rieken, S. Ernst, G. Bagschik, F. Dierkes, M. Nolte, and M. Maurer, "Towards a functional system architecture for automated vehicles," Mar. 24, 2017, *arXiv:1703.08557*.

[4] J. Houston, G. Zuidhof, L. Bergamini, Y. Ye, L. Chen, A. Jain, S. Omari, V. Iglovikov, and P. Ondruska, "One thousand and one hours: self-driving motion prediction dataset," Jun. 25, 2020, *arXiv: 2006.14480*.

[5] "Behind the map: how we keep our maps up to date | TomTom Blog." [Online]. Available: https://www.tomtom.com/blog/maps/continuous-map-processing/

[6] HERE, "Map Data | Static Map API." 2021. [Online]. Available: https://www.here.com/platform/map-data

[7] A. Geiger, P. Lenz, C. Stiller, and R. Urtasun, "Vision meets robotics: The KITTI dataset," *Int. J. Rob. Res.*, vol. 32, no. 11, pp. 1231–1237, Sep. 2013, doi: 10.1177/0278364913491297.

[8] H. Caesar, V. Bankiti, A. H. Lang, S. Vora, V. E. Liong, Q. Xu, A. Krishnan, Y. Pan, G. Baldan, and O. Beijbom, "Nuscenes: A multimodal dataset for autonomous driving," in *Proc. IEEE Comput. Soc. Conf. Comput. Vis. Pattern Recognit.*, 2020, pp. 11618–11628.

[9] T. Shan and B. Englot, "LeGO-LOAM: Lightweight and ground-optimized LiDAR odometry and mapping on variable terrain," in *IEEE Int. Conf. Intell. Robot. Syst.*, 2018, pp. 4758–4765.

[10] W. Xu and F. Zhang, "FAST-LIO: A fast, robust LiDAR-inertial odometry package by tightly-coupled iterated kalman filter," *IEEE Robot. Autom. Lett.*, vol. 6, no. 2, pp. 3317–3324, Apr. 2021, doi: 10.1109/LRA.2021.3064227.

[11] W. Xu, Y. Cai, D. He, J. Lin, and F. Zhang, "FAST-LIO2: Fast direct LiDAR-inertial odometry," *IEEE Trans. Robot.*, Jan. 2022, doi: 10.1109/TRO.2022.3141876.

[12] J. Matsuo, K. Kondo, T. Murakami, T. Sato, Y. Kitsukawa, and J. Meguro, "3D point cloud construction with absolute positions using SLAM based on RTK-GNSS," in *Abstr. Int. Conf. Adv. mechatronics Towar. Evol. fusion IT mechatronics ICAM*, 2021, vol. 2021.7, no. 0, pp. GS7-2.

[13] T. Shan, B. Englot, D. Meyers, W. Wang, C. Ratti, and D. Rus, "LIO-SAM: Tightly-coupled lidar inertial odometry via smoothing and mapping," in *IEEE Int. Conf. Intell. Robot. Syst.*, 2020, pp. 5135–5142.

[14] C. Zheng, Q. Zhu, W. Xu, X. Liu, Q. Guo, and F. Zhang, "FAST-LIVO: Fast and tightly-coupled sparse-direct LiDAR-inertial-visual odometry," Mar. 02, 2022, *arXiv: 2203.00893*.

[15] R. Dubé, A. Cramariuc, D. Dugas, H. Sommer, M. Dymczyk, J. Nieto, R. Siegwart, and C. Cadena, "SegMap: Segment-based mapping and localization using data-driven descriptors," *Int. J. Rob. Res.*, vol. 39, no. 2–3, pp. 339–355, Mar. 2020, doi: 10.1177/0278364919863090.

[16] J. Zhang and S. Singh, "LOAM: Lidar odometry and mapping in real-time," in *Robot. Sci. Syst.*, 2015, vol. 2, no. 9, pp. 1–9.

[17] D. Rozenberszki and A. L. Majdik, "LOL: Lidar-only odometry and localization in 3D point cloud maps∗," in *Proc. IEEE Int. Conf. Robot. Autom.*, 2020, pp. 4379–4385.

[18] J. Lin and F. Zhang, "Loam livox: A fast, robust, high-precision LiDAR odometry and mapping package for LiDARs of small FoV," in *Proc. IEEE Int. Conf. Robot. Autom.*, 2020, pp. 3126–3131.

[19] J. Lin and F. Zhang, "A fast, complete, point cloud based loop closure for LiDAR odometry and mapping," Sep. 25, 2019, *arXiv: 1909.11811*.

[20] J. Lin, X. Liu, and F. Zhang, "A decentralized framework for simultaneous calibration, localization and mapping with multiple LiDARs," in *IEEE Int. Conf. Intell. Robot. Syst.*, 2020, pp. 4870–4877.

[21] B. Fang, J. Ma, P. An, Z. Wang, J. Zhang, and K. Yu, "Multi-level height maps-based registration method for sparse LiDAR point clouds in an urban scene," *Appl. Opt.*, vol. 60, no. 14, p. 4154, May 2021, doi: 10.1364/ao.419746.

[22] A. Gressin, C. Mallet, and N. David, "Improving 3D lidar point cloud registration using optimal neighborhood knowledge," in *ISPRS Ann. Photogramm. Remote Sens. Spat. Inf. Sci.*, 2012, vol. 1, pp. 111–116.

[23] M. Magnusson, A. Lilienthal, and T. Duckett, "Scan registration for autonomous mining vehicles using 3D-NDT," *J. F. Robot.*, vol. 24, no. 10, pp. 803–827, Oct. 2007, doi: 10.1002/rob.20204.

[24] M. Magnusson, "The three-dimensional normal-distributions transform — an efficient representation for registration, surface analysis, and loop detection," Ph.D. dissertation, Dept. of Tech., Örebro universitet, Sweden, 2008.

[25] P. Biber, "The normal distributions transform: A new approach to laser scan matching," in *IEEE Int. Conf. Intell. Robot. Syst.*, 2003, vol. 3, pp. 2743–2748.

[26] E. Takeuchi and T. Tsubouchi, "A 3-D scan matching using improved 3-D normal distributions transform for mobile robotic mapping," in *IEEE Int. Conf. Intell. Robot. Syst.*, 2006, pp. 3068–3073.

[27] S. Kato, S. Tokunaga, Y. Maruyama, S. Maeda, M. Hirabayashi, Y. Kitsukawa, A. Monrroy, T. Ando, Y. Fujii, and T. Azumi, "Autoware on board: Enabling autonomous vehicles with embedded systems," in *Proc. 9th ACM/IEEE Int. Conf. Cyber-Physical Syst. ICCPS 2018*, 2018, pp. 287–296.

[28] K. Chen, R. Nemiroff, and B. T. Lopez, "Direct LiDAR-inertial odometry," Mar. 07, 2022, *arXiv: 2203.03749*.

[29] J. Lin, C. Zheng, W. Xu, and F. Zhang, "R2LIVE: A robust, real-time, LiDAR-inertial-visual tightly-coupled state estimator and mapping," *IEEE Robot. Autom. Lett.*, vol. 6, no. 4, pp. 7469–7476, Jul. 2021, doi: 10.1109/LRA.2021.3095515.

[30] J. Lin and F. Zhang, "R3LIVE: A robust, real-time, RGB-colored, LiDAR-inertial-visual tightly-coupled state estimation and mapping package," Sep. 10, 2021, *arXiv: 2109.07982*.

[31] T. Shan, B. Englot, C. Ratti, and D. Rus, "LVI-SAM: Tightly-coupled Lidar-visual-tnertial odometry via smoothing and mapping," in *Proc. IEEE Int. Conf. Robot. Autom.*, 2021, pp. 5692–5698.

[32] S. Song, S. Jung, H. Kim, and H. Myung, "A method for mapping and localization of quadrotors for inspection under bridges using camera and 3D-LiDAR," in *Proc. 7th Asia-Pacific Work. Struct. Heal. Monit. APWSHM 2018*, 2018, pp. 1061–1068.

[33] "Human in the loop: Machine learning and AI for the people." [Online]. Available: https://www.zdnet.com/article/human-in-the-loop-machine-learning-and-ai-for-the-people/

[34] L. Biewald, "Why human-in-the-loop computing is the future of machine learning | Computerworld," *Computer World*. 2015. [Online]. Available: https://www.computerworld.com/article/3004013/robotics/why-human-in-the-loop-computing-is-the-future-of-machine-learning.html

[35] D. Xin, L. Ma, J. Liu, S. Macke, S. Song, and A. Parameswaran, "Accelerating human-in-the-loop machine learning: Challenges and opportunities," in *Proc. 2nd Work. Data Manag. End-To-End Mach. Learn. DEEM 2018 - conjunction with 2018 ACM SIGMOD/PODS Conf.*, 2018, pp. 1–4.

[36] G. Mattyus, W. Luo, and R. Urtasun, "DeepRoadMapper: Extracting road topology from aerial images," in *Proc. IEEE Int. Conf. Comput. Vis.*, 2017, pp. 3458–3466.

[37] K. He, X. Zhang, S. Ren, and J. Sun, "Deep residual learning for image recognition," in *Proc. IEEE Comput. Soc. Conf. Comput. Vis. Pattern Recognit.*, 2016, pp. 770–778.

[38] T. Y. Zhang and C. Y. Suen, "A fast parallel algorithm for thinning digital patterns," *Commun. ACM*, vol. 27, no. 3, pp. 236–239, Mar. 1984, doi: 10.1145/357994.358023.

[39] P. E. Hart, N. J. Nilsson, and B. Raphael, "A formal basis for the heuristic determination of minimum cost paths," *IEEE Trans. Syst. Sci. Cybern.*, vol. 4, no. 2, pp. 100–107, Jul. 1968, doi: 10.1109/TSSC.1968.300136.

[40] S. Wang, M. Bai, G. Mattyus, H. Chu, W. Luo, B. Yang, J. Liang, J. Cheverie, S. Fidler, and R. Urtasun, "TorontoCity: Seeing the world with a million eyes," in *Proc. IEEE Int. Conf. Comput. Vis.*, 2017, pp. 3028–3036.

[41] J. D. Wegner, J. A. Montoya-Zegarra, and K. Schindler, "Road networks as collections of minimum cost paths," *ISPRS J. Photogramm. Remote Sens.*, vol. 108, pp. 128–137, Oct. 2015, doi: 10.1016/j.isprsjprs.2015.07.002.

[42] A. Batra, S. Singh, G. Pang, S. Basu, C. V. Jawahar, and M. Paluri, "Improved road connectivity by joint learning of orientation and segmentation," in *Proc. IEEE Comput. Soc. Conf. Comput. Vis. Pattern Recognit.*, 2019, pp. 10377–10385.

[43] A. Van Etten, D. Lindenbaum, and T. M. Bacastow, "SpaceNet: A remote sensing dataset and challenge series," Jul. 03, 2018, *arXiv: 1807.01232*.

[44] I. Demir, K. Koperski, D. Lindenbaum, G. Pang, J. Huang, S. Basu, F. Hughes, D. Tuia, and R. Raska, "DeepGlobe 2018: A challenge to parse

the earth through satellite images," in *IEEE Comput. Soc. Conf. Comput. Vis. Pattern Recognit. Work.*, 2018, pp. 172–181.

[45] A. Mosinska, P. Marquez-Neila, M. Kozinski, and P. Fua, "Beyond the pixel-wise loss for topology-aware delineation," in *Proc. IEEE Comput. Soc. Conf. Comput. Vis. Pattern Recognit.*, 2018, pp. 3136–3145.

[46] G. Máttyus and R. Urtasun, "Matching adversarial networks," in *Proc. IEEE Comput. Soc. Conf. Comput. Vis. Pattern Recognit.*, 2018, pp. 8024–8032.

[47] A. Chaurasia and E. Culurciello, "LinkNet: Exploiting encoder representations for efficient semantic segmentation," in *2017 IEEE Vis. Commun. Image Process. VCIP 2017*, 2018, pp. 1–4.

[48] F. Bastani, S. He, S. Abbar, M. Alizadeh, H. Balakrishnan, S. Chawla, S. Madden, and D. Dewitt, "RoadTracer: Automatic extraction of road networks from aerial images," in *Proc. IEEE Comput. Soc. Conf. Comput. Vis. Pattern Recognit.*, 2018, pp. 4720–4728.

[49] H. Ghandorh, W. Boulila, S. Masood, A. Koubaa, F. Ahmed, and J. Ahmad, "Semantic segmentation and edge detection—approach to road detection in very high resolution satellite images," *Remote Sens.*, vol. 14, no. 3, Jan. 2022, doi: 10.3390/rs14030613.

[50] L.-C. Chen, G. Papandreou, F. Schroff, and H. Adam, "Rethinking atrous convolution for semantic image segmentation," 2017, *arXiv: 1706.05587*.

[51] M. Milanova, "Visual attention in deep learning: a review," *Int. Robot. Autom. J.*, vol. 4, no. 3, May 2018, doi: 10.15406/iratj.2018.04.00113.

[52] X. Li, W. Zhang, and Q. Ding, "Understanding and improving deep learning-based rolling bearing fault diagnosis with attention mechanism," *Signal Processing*, vol. 161, pp. 136–154, Aug. 2019, doi: 10.1016/j.sigpro.2019.03.019.

[53] Y. Chen, G. Peng, Z. Zhu, and S. Li, "A novel deep learning method based on attention mechanism for bearing remaining useful life prediction," *Appl. Soft Comput. J.*, vol. 86, Jan. 2020, doi: 10.1016/j.asoc.2019.105919.

[54] Z. Niu, G. Zhong, and H. Yu, "A review on the attention mechanism of deep learning," *Neurocomputing*, vol. 452, pp. 48–62, Sep. 2021, doi: 10.1016/j.neucom.2021.03.091.

[55] G. Cheng, Y. Wang, S. Xu, H. Wang, S. Xiang, and C. Pan, "Automatic road detection and centerline extraction via cascaded end-to-end convolutional neural network," *IEEE Trans. Geosci. Remote Sens.*, vol. 55, no. 6, pp. 3322–3337, Mar. 2017, doi: 10.1109/TGRS.2017.2669341.

[56] N. Xue, S. Bai, F. Wang, G. S. Xia, T. Wu, and L. Zhang, "Learning attraction field representation for robust line segment detection," in *Proc. IEEE Comput. Soc. Conf. Comput. Vis. Pattern Recognit.*, 2019, pp. 1595–1603.

[57] S. He, F. Bastani, S. Jagwani, M. Alizadeh, H. Balakrishnan, S. Chawla, M. M. Elshrif, S. Madden, and M. A. Sadeghi, "Sat2Graph: Road graph extraction through graph-tensor encoding," in *Lect. Notes Comput. Sci. (including Subser. Lect. Notes Artif. Intell. Lect. Notes Bioinformatics)*, 2020, vol. 12369 LNCS, pp. 51–67.

[58] N. Girard, D. Smirnov, J. Solomon, and Y. Tarabalka, "Polygonal building extraction by frame field learning," in *Proc. IEEE Comput. Soc. Conf. Comput. Vis. Pattern Recognit.*, 2021, pp. 5887–5896.

[59] A. Vaswani, N. Shazeer, N. Parmar, J. Uszkoreit, L. Jones, A. N. Gomez, Ł. Kaiser, and I. Polosukhin, "Attention is all you need," in *Adv. Neural Inf. Process. Syst.*, 2017, pp. 5999–6009.

[60] Z. Xu, Y. Liu, L. Gan, X. Hu, Y. Sun, M. Liu, and L. Wang, "CsBoundary: City-scale road-boundary detection in aerial images for high-definition maps," *IEEE Robot. Autom. Lett.*, vol. 7, no. 2, pp. 5063–5070, Apr. 2022, doi: 10.1109/LRA.2022.3154052.

[61] T. Y. Lin, P. Dollár, R. Girshick, K. He, B. Hariharan, and S. Belongie, "Feature pyramid networks for object detection," in *Proc. 30th IEEE Conf. Comput. Vis. Pattern Recognition, CVPR 2017*, 2017, pp. 936–944.

[62] Z. Xu, Y. Sun, and M. Liu, "Topo-Boundary: A benchmark dataset on topological road-boundary detection using aerial images for autonomous driving," *IEEE Robot. Autom. Lett.*, vol. 6, no. 4, pp. 7248–7255, Jul. 2021, doi: 10.1109/LRA.2021.3097512.

[63] J. D. Wegner, J. A. Montoya-Zegarra, and K. Schindler, "A higher-order CRF model for road network extraction," in *Proc. IEEE Comput. Soc. Conf. Comput. Vis. Pattern Recognit.*, 2013, pp. 1698–1705.

[64] M. Ibrahim, N. Akhtar, M. A. A. K. Jalwana, M. Wise, and A. Mian, "High Definition LiDAR mapping of Perth CBD," in *2021 Digit. Image Comput. Tech. Appl.*, 2021, pp. 1–8.

[65] D. Landry, F. Pomerleau, and P. Giguère, "CELLO-3D: Estimating the covariance of ICP in the real world," in *Proc. IEEE Int. Conf. Robot.*

*Autom.*, 2019, pp. 8190–8196.

[66] I. Bogoslavskyi and C. Stachniss, "Efficient online segmentation for sparse 3D laser scans," *Photogramm. Fernerkundung, Geoinf.*, vol. 85, no. 1, pp. 41–52, Feb. 2017, doi: 10.1007/s41064-016-0003-y.

[67] S. Ding, Y. Fu, W. Wang, and Z. Pan, "Creation of high definition map for autonomous driving within specific scene," in *Int. Conf. Smart Transp. City Eng. 2021*, 2021, pp. 1365–1373.

[68] S. Gu, Y. Zhang, J. Yang, J. M. Alvarez, and H. Kong, "Two-view fusion based convolutional neural network for urban road detection," in *IEEE Int. Conf. Intell. Robot. Syst.*, 2019, pp. 6144–6149.

[69] S. Gu, Y. Zhang, J. Tang, J. Yang, and H. Kong, "Road detection through CRF based LiDAR-camera fusion," in *Proc. IEEE Int. Conf. Robot. Autom.*, 2019, pp. 3832–3838.

[70] D. Yu, H. Xiong, Q. Xu, J. Wang, and K. Li, "Multi-stage residual fusion network for LIDAR-camera road detection," in *IEEE Intell. Veh. Symp. Proc.*, 2019, pp. 2323–2328.

[71] Y. Li, L. Xiang, C. Zhang, and H. Wu, "Fusing taxi trajectories and rs images to build road map via dcnn," *IEEE Access*, vol. 7, pp. 161487–161498, Nov. 2019, doi: 10.1109/ACCESS.2019.2951730.

[72] B. Li, D. Song, A. Ramchandani, H. M. Cheng, D. Wang, Y. Xu, and B. Chen, "Virtual lane boundary generation for human-compatible autonomous driving: A tight coupling between perception and planning," in *IEEE Int. Conf. Intell. Robot. Syst.*, 2019, pp. 3733–3739.

[73] L. Ma, Y. Li, J. Li, J. M. Junior, W. N. Goncalves, and M. A. Chapman, "BoundaryNet: Extraction and completion of road boundaries with deep learning using mobile laser scanning point clouds and satellite imagery," *IEEE Trans. Intell. Transp. Syst.*, Feb. 2021, doi: 10.1109/TITS.2021.3055366.

[74] O. Ronneberger, P. Fischer, and T. Brox, "U-net: Convolutional networks for biomedical image segmentation," in *Lect. Notes Comput. Sci. (including Subser. Lect. Notes Artif. Intell. Lect. Notes Bioinformatics)*, 2015, vol. 9351, pp. 234–241.

[75] I. Goodfellow, J. Pouget-Abadie, M. Mirza, B. Xu, D. Warde-Farley, S. Ozair, A. Courville, and Y. Bengio, "Generative adversarial networks," *Commun. ACM*, vol. 63, no. 11, pp. 139–144, Nov. 2020, doi: 10.1145/3422622.

[76] L. Zhou, C. Zhang, and M. Wu, "D-linknet: Linknet with pretrained encoder and dilated convolution for high resolution satellite imagery road extraction," in *IEEE Comput. Soc. Conf. Comput. Vis. Pattern Recognit. Work.*, 2018, pp. 192–196.

[77] M. Schreiber, C. Knöppel, and U. Franke, "LaneLoc: Lane marking based localization using highly accurate maps," in *IEEE Intell. Veh. Symp. Proc.*, 2013, pp. 449–454.

[78] W. Jang, J. An, S. Lee, M. Cho, M. Sun, and E. Kim, "Road lane semantic segmentation for high definition map," in *IEEE Intell. Veh. Symp. Proc.*, 2018, pp. 1001–1006.

[79] J. Jiao, "Machine learning assisted high-definition map creation," in *Proc. Int. Comput. Softw. Appl. Conf.*, 2018, vol. 1, pp. 367–373.

[80] M. Aldibaja, N. Suganuma, and K. Yoneda, "LIDAR-data accumulation strategy to generate high definition maps for autonomous vehicles," in *IEEE Int. Conf. Multisens. Fusion Integr. Intell. Syst.*, 2017, pp. 422–428.

[81] X. Mi, B. Yang, Z. Dong, C. Liu, Z. Zong, and Z. Yuan, "A two-stage approach for road marking extraction and modeling using MLS point clouds," *ISPRS J. Photogramm. Remote Sens.*, vol. 180, pp. 255–268, Oct. 2021, doi: 10.1016/j.isprsjprs.2021.07.012.

[82] D. Betaille and R. Toledo-Moreo, "Creating enhanced maps for lane-level vehicle navigation," *IEEE Trans. Intell. Transp. Syst.*, vol. 11, no. 4, pp. 786–798, Jul. 2010, doi: 10.1109/TITS.2010.2050689.

[83] Z. Zhang, X. Zhang, S. Yang, and J. Yang, "Research on pavement marking recognition and extraction method," in *2021 6th Int. Conf. Image, Vis. Comput. ICIVC 2021*, 2021, pp. 100–105.

[84] Y. Zhang, Z. Lu, D. Ma, J. H. Xue, and Q. Liao, "Ripple-GAN: Lane line detection with ripple lane line detection network and wasserstein GAN," *IEEE Trans. Intell. Transp. Syst.*, vol. 22, no. 3, pp. 1532–1542, May 2021, doi: 10.1109/TITS.2020.2971728.

[85] N. Kanopoulos, N. Vasanthavada, and R. L. Baker, "Design of an image edge detection filter using the Sobel operator," *IEEE J. Solid-State Circuits*, vol. 23, no. 2, pp. 358–367, Apr. 1988, doi: 10.1109/4.996.

[86] M. Arjovsky, S. Chintala, and L. Bottou, "Wasserstein GAN," Jan. 26, 2017, *arXiv: 1701.07875*.

[87] J. Zhang, T. Deng, F. Yan, and W. Liu, "Lane detection model based on spatio-temporal network with double convolutional gated recurrent units," *IEEE Trans. Intell. Transp. Syst.*, Feb. 2021, doi: 10.1109/TITS.2021.3060258.

[88] M. Siam, S. Valipour, M. Jagersand, and N. Ray, "Convolutional gated recurrent networks for video segmentation," in *Proc. Int. Conf. Image Process. ICIP*, 2018, pp. 3090–3094. doi: 10.1109/ICIP.2017.8296851.

[89] B. He, Y. Guan, and R. Dai, "Convolutional gated recurrent units for medical relation classification," in *Proc. 2018 IEEE Int. Conf. Bioinforma. Biomed. BIBM 2018*, 2019, pp. 646–650.

[90] P. Lu, S. Xu, and H. Peng, "Graph-embedded lane detection," *IEEE Trans. Image Process.*, vol. 30, pp. 2977–2988, Feb. 2021, doi: 10.1109/TIP.2021.3057287.

[91] M. Marzougui, A. Alasiry, Y. Kortli, and J. Baili, "A lane tracking method based on progressive probabilistic hough transform," *IEEE Access*, vol. 8, pp. 84893–84905, May 2020, doi: 10.1109/ACCESS.2020.2991930.

[92] X. Xu, T. Yu, X. Hu, W. W. Y. Ng, and P. A. Heng, "SALMNet: A structure-aware lane marking detection network," *IEEE Trans. Intell. Transp. Syst.*, vol. 22, no. 8, pp. 4986–4997, Apr. 2021, doi: 10.1109/TITS.2020.2983017.

[93] Y. Ko, Y. Lee, S. Azam, F. Munir, M. Jeon, and W. Pedrycz, "Key points estimation and point instance segmentation approach for lane detection," *IEEE Trans. Intell. Transp. Syst.*, Jun. 2021, doi: 10.1109/TITS.2021.3088488.

[94] J. Tian, J. Yuan, and H. Liu, "Road marking detection based on mask R-CNN instance segmentation model," in *Proc. 2020 Int. Conf. Comput. Vision, Image Deep Learn. CVIDL 2020*, 2020, pp. 246–249.

[95] "GitHub - TuSimple/tusimple-benchmark: Download datasets and ground truths." [Online]. Available: https://github.com/TuSimple/tusimple-benchmark

[96] X. Pan, J. Shi, P. Luo, X. Wang, and X. Tang, "Spatial as deep: Spatial CNN for traffic scene understanding," in *32nd AAAI Conf. Artif. Intell. AAAI 2018*, 2018, pp. 7276–7283.

[97] D. Neven, B. De Brabandere, S. Georgoulis, M. Proesmans, and L. Van Gool, "Towards end-to-end lane detection: An instance segmentation approach," in *IEEE Intell. Veh. Symp. Proc.*, 2018, pp. 286–291.

[98] X. Li, J. Li, X. Hu, and J. Yang, "Line-CNN: End-to-end traffic line detection with line proposal unit," *IEEE Trans. Intell. Transp. Syst.*, vol. 21, no. 1, pp. 248–258, Jan. 2020, doi: 10.1109/TITS.2019.2890870.

[99] S. M. Azimi, P. Fischer, M. Korner, and P. Reinartz, "Aerial LaneNet: Lane-marking semantic segmentation in aerial imagery using wavelet-enhanced cost-sensitive symmetric fully convolutional neural networks," *IEEE Trans. Geosci. Remote Sens.*, vol. 57, no. 5, pp. 2920–2938, Mar. 2019, doi: 10.1109/TGRS.2018.2878510.

[100] F. Kurz, S. M. Azimi, C. Y. Sheu, and P. D'Angelo, "Deep learning segmentation and 3D reconstruction of road markings using multiview aerial imagery," *ISPRS Int. J. Geo-Information*, vol. 8, no. 1, Jan. 2019, doi: 10.3390/ijgi8010047.

[101] Y. Yu, Y. Li, C. Liu, J. Wang, C. Yu, X. Jiang, L. Wang, Z. Liu, and Y. Zhang, "MarkCapsNet: Road marking extraction from aerial images using self-attention-guided capsule network," *IEEE Geosci. Remote Sens. Lett.*, vol. 19, Nov. 2022, doi: 10.1109/LGRS.2021.3124575.

[102] J. Wang, K. Sun, T. Cheng, B. Jiang, C. Deng, Y. Zhao, D. Liu, Y. Mu, M. Tan, X. Wang, W. Liu, and B. Xiao, "Deep high-resolution representation learning for visual recognition," *IEEE Trans. Pattern Anal. Mach. Intell.*, vol. 43, no. 10, pp. 3349–3364, Oct. 2021, doi: 10.1109/TPAMI.2020.2983686.

[103] C. Wen, X. Sun, J. Li, C. Wang, Y. Guo, and A. Habib, "A deep learning framework for road marking extraction, classification and completion from mobile laser scanning point clouds," *ISPRS J. Photogramm. Remote Sens.*, vol. 147, pp. 178–192, Jan. 2019, doi: 10.1016/j.isprsjprs.2018.10.007.

[104] R. LaLonde, Z. Xu, I. Irmakci, S. Jain, and U. Bagci, "Capsules for biomedical image segmentation," *Med. Image Anal.*, vol. 68, Feb. 2021, doi: 10.1016/j.media.2020.101889.

[105] Y. Yu, Y. Yao, H. Guan, D. Li, Z. Liu, L. Wang, C. Yu, S. Xiao, W. Wang, and L. Chang, "A self-attention capsule feature pyramid network for water body extraction from remote sensing imagery," *Int. J. Remote Sens.*, vol. 42, no. 5, pp. 1801–1822, Dec. 2021, doi: 10.1080/01431161.2020.1842544.

[106] C. Ye, J. Li, H. Jiang, H. Zhao, L. Ma, and M. Chapman, "Semi-automated generation of road transition lines using mobile laser scanning data," *IEEE Trans. Intell. Transp. Syst.*, vol. 21, no. 5, pp. 1877–1890, May 2020, doi: 10.1109/TITS.2019.2904735.

[107] L. Ma, Y. Li, J. Li, Z. Zhong, and M. A. Chapman, "Generation of horizontally curved driving lines in HD maps using mobile laser scanning point clouds," *IEEE J. Sel. Top. Appl. Earth Obs. Remote Sens.*, vol. 12, no. 5, pp. 1572–1586, May 2019, doi: 10.1109/JSTARS.2019.2904514.

[108] M. Soilán, B. Riveiro, J. Martínez-Sánchez, and P. Arias, "Segmentation and classification of road markings using MLS data," *ISPRS J. Photogramm. Remote Sens.*, vol. 123, pp. 94–103, Jan. 2017, doi: 10.1016/j.isprsjprs.2016.11.011.

[109] M. Soilán, A. Sánchez-Rodríguez, P. Del Río-Barral, C. Perez-Collazo, P. Arias, and B. Riveiro, "Review of laser scanning technologies and their applications for road and railway infrastructure monitoring," *Infrastructures*, vol. 4, no. 4, Sep. 2019, doi: 10.3390/infrastructures4040058.

[110] B. He, R. Ai, Y. Yan, and X. Lang, "Lane marking detection based on convolution neural network from point clouds," in *IEEE Conf. Intell. Transp. Syst. Proceedings, ITSC*, 2016, pp. 2475–2480.

[111] L. Ma, Y. Li, J. Li, Y. Yu, J. M. Junior, W. N. Goncalves, and M. A. Chapman, "Capsule-based networks for road marking extraction and classification from mobile LiDAR point clouds," *IEEE Trans. Intell. Transp. Syst.*, vol. 22, no. 4, pp. 1981–1995, Apr. 2021, doi: 10.1109/TITS.2020.2990120.

[112] S. Sabour, N. Frosst, and G. E. Hinton, "Dynamic routing between capsules," in *Adv. Neural Inf. Process. Syst.*, 2017, pp. 3857–3867.

[113] M. Cheng, H. Zhang, C. Wang, and J. Li, "Extraction and classification of road markings using mobile laser scanning point clouds," *IEEE J. Sel. Top. Appl. Earth Obs. Remote Sens.*, vol. 10, no. 3, pp. 1182–1196, Mar. 2017, doi: 10.1109/JSTARS.2016.2606507.

[114] D. Prochazka, J. Prochazkova, and J. Landa, "Automatic lane marking extraction from point cloud into polygon map layer," *Eur. J. Remote Sens.*, vol. 52, no. sup1, pp. 26–39, Oct. 2019, doi: 10.1080/22797254.2018.1535837.

[115] J. Landa, D. Prochazka, and J. Šťastny, "Point cloud processing for smart systems," *Acta Univ. Agric. Silvic. Mendelianae Brun.*, vol. 61, no. 7, pp. 2415–2421, Jan. 2013, doi: 10.11118/actaun201361072415.

[116] J. Redmon and A. Farhadi, "YOLOv3: An incremental improvement," Apr. 08, 2018, *arXiv: 1804.02767*.

[117] M. Lehtomäki, A. Jaakkola, J. Hyyppä, A. Kukko, and H. Kaartinen, "Detection of vertical pole-like objects in a road environment using vehicle-based laser scanning data," *Remote Sens.*, vol. 2, no. 3, pp. 641–664, Mar. 2010, doi: 10.3390/rs2030641.

[118] S. I. El-Halawany and D. D. Lichti, "Detection of road poles from mobile terrestrial laser scanner point cloud," in *2011 Int. Work. Multi-Platform/Multi-Sensor Remote Sens. Mapping, M2RSM 2011*, 2011, pp. 1–6.

[119] H. Yokoyama, H. Date, S. Kanai, and H. Takeda, "Pole-like objects recognition from mobile laser scanning data using smoothing and principal component analysis," *Int. Arch. Photogramm. Remote Sens. Spat. Inf. Sci.*, vol. XXXVIII–5, pp. 115–120, Sep. 2012, doi: 10.5194/isprsarchives-xxxviii-5-w12-115-2011.

[120] S. Pu, M. Rutzinger, G. Vosselman, and S. Oude Elberink, "Recognizing basic structures from mobile laser scanning data for road inventory studies," *ISPRS J. Photogramm. Remote Sens.*, vol. 66, no. 6 SUPPL., Dec. 2011, doi: 10.1016/j.isprsjprs.2011.08.006.

[121] C. Cabo, C. Ordóñez, S. García-Cortés, and J. Martínez, "An algorithm for automatic detection of pole-like street furniture objects from Mobile Laser Scanner point clouds," *ISPRS J. Photogramm. Remote Sens.*, vol. 87, pp. 47–56, Jan. 2014, doi: 10.1016/j.isprsjprs.2013.10.008.

[122] C. Ordóñez, C. Cabo, and E. Sanz-Ablanedo, "Automatic detection and classification of pole-like objects for urban cartography using mobile laser scanning data," *Sensors (Switzerland)*, vol. 17, no. 7, Jun. 2017, doi: 10.3390/s17071465.

[123] Y. Yu, J. Li, H. Guan, C. Wang, and J. Yu, "Semiautomated extraction of street light poles from mobile LiDAR point-clouds," *IEEE Trans. Geosci. Remote Sens.*, vol. 53, no. 3, pp. 1374–1386, Mar. 2015, doi: 10.1109/TGRS.2014.2338915.

[124] H. Zheng, R. Wang, and S. Xu, "Recognizing street lighting poles from mobile LiDAR data," *IEEE Trans. Geosci. Remote Sens.*, vol. 55, no. 1, pp. 407–420, Jan. 2017, doi: 10.1109/TGRS.2016.2607521.

[125] C. Plachetka, J. Fricke, M. Klingner, and T. Fingscheidt, "DNN-based recognition of pole-like objects in LiDAR point clouds," in *IEEE Conf. Intell. Transp. Syst. Proceedings, ITSC*, 2021, pp. 2889–2896.

[126] C. R. Qi, H. Su, K. Mo, and L. J. Guibas, "PointNet: Deep learning on point sets for 3D classification and segmentation," in *Proc. - 30th IEEE Conf. Comput. Vis. Pattern Recognition, CVPR 2017*, 2017, pp. 77–85.

[127] A. H. Lang, S. Vora, H. Caesar, L. Zhou, J. Yang, and O. Beijbom, "Pointpillars: Fast encoders for object detection from point clouds," in *Proc. IEEE Comput. Soc. Conf. Comput. Vis. Pattern Recognit.*, 2019, pp. 12689–12697.

[128] Y. Zhou and O. Tuzel, "VoxelNet: End-to-end learning for point cloud based 3D object detection," in *Proc. IEEE Comput. Soc. Conf. Comput. Vis. Pattern Recognit.*, 2018, pp. 4490–4499.
[129] P. Hu and D. Ramanan, "Finding tiny faces," in *Proc. 30th IEEE Conf. Comput. Vis. Pattern Recognition, CVPR 2017*, 2017, pp. 1522–1530.
[130] W. Liu, D. Anguelov, D. Erhan, C. Szegedy, S. Reed, C. Y. Fu, and A. C. Berg, "SSD: Single shot multibox detector," in *Lect. Notes Comput. Sci. (including Subser. Lect. Notes Artif. Intell. Lect. Notes Bioinformatics)*, 2016, vol. 9905 LNCS, pp. 21–37.
[131] F. Poggenhans, J. H. Pauls, J. Janosovits, S. Orf, M. Naumann, F. Kuhnt, and M. Mayr, "Lanelet2: A high-definition map framework for the future of automated driving," in *IEEE Conf. Intell. Transp. Syst. Proceedings, ITSC*, 2018, pp. 1672–1679.
[132] "ASAM OpenDRIVE®." [Online]. Available: https://www.asam.net/standards/detail/opendrive/
[133] C. W. Gran, "HD-maps in autonomous driving," M.S. thesis, Dept. Comp. Sci., Norwegian Univ. of Sci. and Tech., Norway, 2019.
[134] "Lanelet2 map for Autoware.Auto." [Online]. Available: https://autowarefoundation.gitlab.io/autoware.auto/AutowareAuto/lanelet2-map-for-autoware-auto.html
[135] "OpenDRIVE 1.6.1." [Online]. Available: https://www.asam.net/index.php?eID=dumpFile&t=f&f=4089&token=deea5d707e2d0edeeb4fccd544a973de4bc46a09#_specific_lane_rules
[136] A. Avila, "Using open source frameworks in autonomous vehicle development - Part 2." Mar. 04, 2020. [Online]. Available: https://auro.ai/blog/2020/03/using-open-source-frameworks-in-autonomous-vehicle-development-part-2/
[137] "Generate maps with OpenStreetMap - CARLA Simulator." [Online]. Available: https://carla.readthedocs.io/en/latest/tuto_G_openstreetmap/
[138] K. Kim, S. Cho, and W. Chung, "HD map update for autonomous driving with crowdsourced data," *IEEE Robot. Autom. Lett.*, vol. 6, no. 2, pp. 1895–1901, Apr. 2021, doi: 10.1109/LRA.2021.3060406.